\documentclass[sigconf]{acmart}

\AtBeginDocument{%
  }

\usepackage{enumitem}
\usepackage{xcolor}

\usepackage{multirow}
\usepackage{booktabs}  
\usepackage{pifont}     
\usepackage{graphicx}  
\usepackage{array}
\usepackage{makecell}
\usepackage{subcaption}
\usepackage{xcolor} 
\usepackage{algorithm}

\usepackage{algpseudocode} 

\usepackage[most]{tcolorbox}
\newtcolorbox{keydiscovery1}{
  colback=gray!5,
  colframe=gray!50,
  boxrule=0.6pt,
  arc=2pt,
  left=6pt,
  right=6pt,
  top=6pt,
  bottom=6pt,
  before skip=6pt,
  after skip=8pt
}

\newtcolorbox{insightbox}{
  colback=gray!3,
  colframe=gray!35,
  boxrule=0.5pt,
  arc=2pt,
  left=6pt,
  right=6pt,
  top=6pt,
  bottom=6pt,
  before skip=6pt,
  after skip=8pt
}

\setcopyright{acmlicensed}
\copyrightyear{2018}
\acmYear{2018}
\acmDOI{XXXXXXX.XXXXXXX}

\acmConference[Conference acronym 'XX]{Make sure to enter the correct
  conference title from your rights confirmation email}{June 03--05,
  2018}{Woodstock, NY}

\acmISBN{978-1-4503-XXXX-X/2018/06}

\begin{document}

\title{TemporalBench: A  Benchmark for Evaluating LLM-Based Agents on Contextual and Event-Informed Time Series Tasks}

\author{Muyan Weng}
\email{muyanwen@usc.edu}
\orcid{0009-0005-8723-0689}
\affiliation{%
  \institution{University of Southern California}
  \city{Los Angeles}
  \state{California}
  \country{USA}
}

\author{Defu Cao}
\email{defucao@usc.edu}
\orcid{0000-0003-0240-3818}
\affiliation{%
  \institution{University of Southern California}
  \city{Los Angeles}
  \state{California}
  \country{USA}
}

\author{Wei Yang}
\email{wyang930@usc.edu}
\orcid{0009-0007-0986-9617}
\affiliation{%
  \institution{University of Southern California}
  \city{Los Angeles}
  \state{California}
  \country{USA}
}

\author{Yashaswi Sharma}
\email{yashaswi@usc.edu}
\orcid{0009-0000-4068-7992}
\affiliation{%
  \institution{University of Southern California}
  \city{Los Angeles}
  \state{California}
  \country{USA}
}

\author{Yan Liu}
\email{yanliu.cs@usc.edu}
\orcid{0000-0002-7055-9518}
\affiliation{%
  \institution{University of Southern California}
  \city{Los Angeles}
  \state{California}
  \country{USA}
}

\newcommand{\modelname}{TemporalBench}

\begin{abstract}
Large language model (LLM)–based agents are increasingly applied to time-series forecasting tasks, yet existing evaluations primarily emphasize numerical accuracy, making it unclear whether strong forecasting performance reflects genuine temporal understanding or the ability to reason under contextual and event-driven conditions.
We introduce \textbf{TemporalBench}, a multi-domain benchmark designed to evaluate temporal reasoning behavior under progressively richer informational settings.
TemporalBench adopts a four-tier task taxonomy that examines historical structure interpretation, context-free forecasting, contextual temporal reasoning, and event-conditioned prediction across four real-world domains: retail, healthcare, energy, and physical systems.
By controlling access to future targets and contextual information, the benchmark enables a diagnostic analysis of whether models can correctly interpret temporal patterns, align them with external context, and adapt predictions when conditions change.
Extensive baseline experiments show that strong numerical forecasting accuracy does not reliably translate into robust contextual or event-aware temporal reasoning; instead, existing agent frameworks exhibit fragmented strengths and systematic failure modes that remain largely hidden under forecasting-only benchmarks.
The TemporalBench dataset is publicly available at \url{https://huggingface.co/datasets/Melady/TemporalBench}, and we additionally provide a public leaderboard at \url{https://huggingface.co/spaces/Melady/TemporalBench_Leaderboard}.

\end{abstract}






\newcommand{\df}[1]{{\color{red}DF: #1}}
\maketitle

\section{Introduction}
Time-series data from real-world systems are rarely generated by a single, stationary process.
Instead, they arise from evolving environments in which external conditions, interventions, and unexpected events repeatedly perturb underlying dynamics.\cite{wu2025out,menchetti2021estimating}
Across domains such as retail, healthcare and energy, temporal patterns are shaped by contextual factors that influence how historical behavior should be interpreted and how future outcomes should be anticipated.
Consequently, effective time-series forecasting in practice requires reasoning about temporal signals under changing conditions, rather than simple extrapolation of past observations.

Despite this inherent complexity, much of the existing time-series evaluation landscape focuses on simplified settings.
To enable controlled comparison and scalable benchmarking, widely used datasets often abstract away contextual variation and event-driven regime changes, emphasizing numerical prediction accuracy under relatively stable assumptions.\cite{shchur2025fev,chang2025time,qiu2024tfb}
While this paradigm has driven substantial progress in forecasting and foundation models~\cite{cao2025timedit, ansari2024chronos, yang2025foundation, cao2023tempo}, it also raises a key concern:
strong performance under simplified benchmarks may not reflect temporal understanding that transfers to real-world scenarios.
In particular, it remains unclear whether forecasting accuracy reflects genuine temporal understanding or merely effective numerical extrapolation.

Recent advances in large language models (LLMs) and agent-based systems further amplify this tension.
LLM-based agents have demonstrated strong capabilities across a range of complex tasks, including software engineering, scientific discovery, robotics planning, and data analysis~\cite{plaat2025agentic, jimenez2023swe, boiko2023emergent, singh2024malmm, chen2025tourrank}.
By combining instruction following, multi-step reasoning, tool use, and access to external knowledge, agent frameworks offer a flexible paradigm for tackling problems that extend beyond single-turn prediction~\cite{ping2025hdlcore,ping2025verimoa,yang2025maestro,yang2025learning,cao2025conversational}.
Motivated by these successes, recent work has begun to explore whether agentic approaches can also benefit time-series analysis and forecasting, particularly in settings involving heterogeneous information and more complex decision logic~\cite{zhao2025timeseriesscientist, zhang2024large, yeh2025empowering,ye2025llm,ye2024domain}.

This line of work implicitly assumes that agent-based systems can move beyond numerical prediction to reason about time-series behavior under context.
However, existing evaluation protocols make this assumption difficult to verify.
Forecasting benchmarks primarily assess numerical prediction error, without evaluating whether models correctly interpret historical structure, align contextual information with temporal segments, or adapt predictions when conditions change~\cite{qiu2024tfb, aksu2024gifteval}.
Conversely, temporal reasoning benchmarks for language models typically operate on textual data alone, omitting real numerical time-series signals and the dynamics they encode~\cite{ning2020torque, fatemi2024test, wang2024tram}.
As a result, it remains unclear whether agent-based systems genuinely improve contextual temporal reasoning, or whether apparent gains instead reflect evaluation protocols that conflate numerical accuracy with broader reasoning capabilities.

A central challenge in bridging this gap lies in the role of \emph{context}.
In real-world time-series applications, contextual information does not merely refine numerical estimates; it shapes how temporal patterns should be interpreted.
Context determines which temporal segments are comparable, which changes are meaningful, and how historical behavior should generalize to the future.
Events such as promotions, policy changes, medical interventions, or extreme weather can induce regime shifts that invalidate naive extrapolation~\cite{williams2024context, wang2024news}.
Evaluating temporal reasoning in such settings therefore requires benchmarks that incorporate context and events as first-class components, rather than treating them as auxiliary inputs.

To address these challenges, we introduce \textbf{TemporalBench}, a multi-domain benchmark for evaluating agent and LLM performance on time-series tasks under progressively richer contextual and event-driven settings.
TemporalBench spans four real-world domains---retail, healthcare, energy, and physical systems---and adopts a diagnostic task-centric design with a four-tier taxonomy (T1--T4) that disentangles historical understanding, context-free prediction, contextual reasoning, and event-informed forecasting.
By controlling access to future targets and contextual information, the benchmark enables systematic analysis of how temporal signals are interpreted and reused.
Through extensive baseline evaluation, we show that strong forecasting accuracy does not reliably translate into robust contextual or event-aware temporal reasoning.

Overall, our contributions are as follows:
\begin{itemize}
\item We introduce \textbf{TemporalBench}, a multi-domain benchmark for evaluating agent and LLM performance on time-series tasks under contextual and event-informed settings.
\item We propose a four-tier task taxonomy (T1--T4) that disentangles historical understanding, pure temporal prediction, contextual reasoning, and event-conditioned forecasting.
\item We develop a unified benchmark construction pipeline with automatic task generation, rule-based ground-truth labeling, and explicit uncertainty modeling.
\item Through extensive baseline evaluation, we demonstrate that strong forecasting accuracy does not reliably translate into robust contextual or event-aware temporal reasoning.
\end{itemize}

\section{TemporalBench}
\label{sec:benchmark}

\subsection{Benchmark Overview}
\begin{table}[htbp]
\centering
\setlength{\tabcolsep}{4pt}
\renewcommand{\arraystretch}{1.1}
\resizebox{\columnwidth}{!}{
\begin{tabular}{lccccc}
\toprule
\textbf{Benchmark} 
& \textbf{TS-Involved} 
& \textbf{Context} 
& \textbf{Reasoning} 
& \textbf{\#Tasks} 
& \textbf{Task Type} \\
\midrule
Test of Time~\cite{fatemi2024test}            
& \ding{55} & \ding{55} & \ding{51} & 1 & QA \\
TRAM~\cite{wang2024tram}                      
& \ding{55} & \ding{55} & \ding{51} & 1 & QA \\
\midrule
TSI-Bench~\cite{du2024tsi}                    
& \ding{51} & \ding{55} & \ding{55} & 1 & TS Analysis \\
TSB-AD~\cite{liu2024elephant}                      
& \ding{51} & \ding{55} & \ding{55} & 1 & TS Analysis \\
GIFT-Eval~\cite{aksu2024gifteval}                     
& \ding{51} & \ding{55} & \ding{55} & 1 & TS Analysis \\
TFB~\cite{qiu2024tfb}                         
& \ding{51} & \ding{55} & \ding{55} & 1 & TS Analysis \\
Time-MMD~\cite{liu2024time}                
& \ding{51} & \ding{55} & \ding{55} & 1 & TS Analysis \\
\midrule
CiK~\cite{williams2024context}                
& \ding{51} & \ding{51} & \ding{55} & 1 & TS Analysis \\
TGTSF~\cite{xu2024intervention}                     
& \ding{51} & \ding{51} & \ding{55} & 1 & TS Analysis \\
\midrule
LLM TS Struggle~\cite{merrill2024language}    
& \ding{51} & \ding{55} & \ding{51} & 2 & QA, TS Analysis \\
MTBench~\cite{chen2025mtbench}                
& \ding{51} & \ding{55} & \ding{51} & 3 & QA, TS Analysis \\
ChatTime~\cite{wang2025chattime}                       
& \ding{51} & \ding{55} & \ding{51} & 3 & QA, TS Analysis \\
\midrule
\textbf{\modelname (Ours)}                    
& \ding{51} & \ding{51} & \ding{51} & 4 & QA, TS Analysis \\
\bottomrule
\end{tabular}
}

\caption{
Comparison of TemporalBench  with existing temporal and time-series related benchmarks.
TS-Involved indicates whether real numerical time-series are included.
Context indicates whether external contextual information is explicitly incorporated.
}

\vspace{-0.8cm}
\label{tab:ts-benchmarks-comparison}
\end{table}

\begin{figure*}[t]
    \centering
\includegraphics[width=\textwidth]{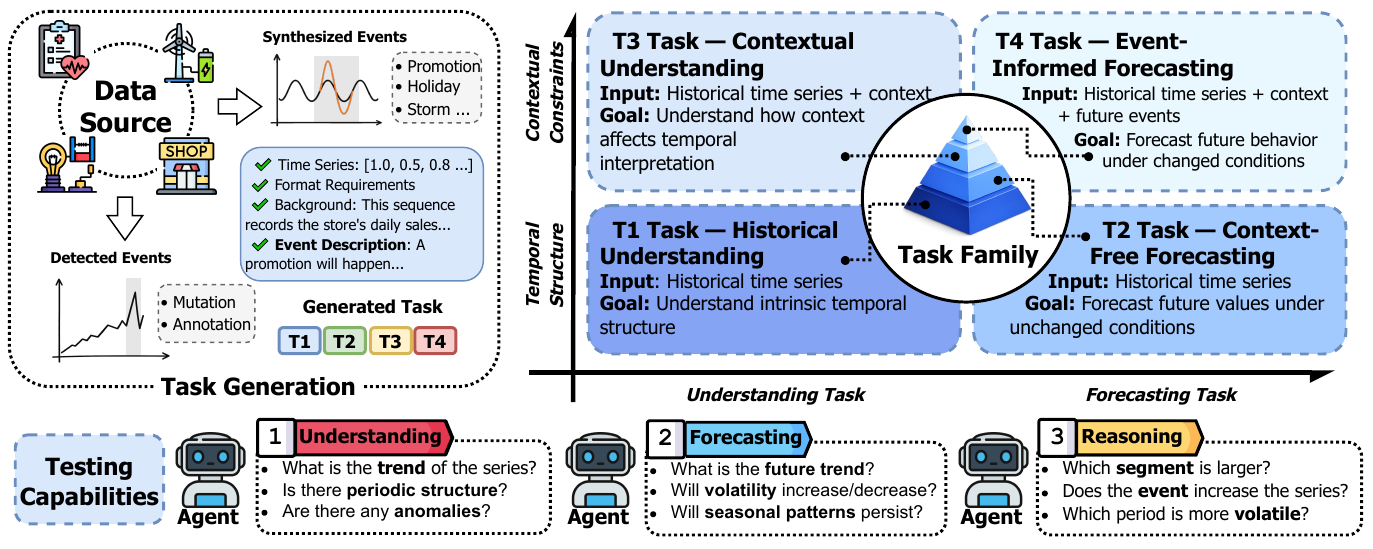}
\vspace{-20pt}
\caption{
Overview of the task generation pipeline in our benchmark, illustrating how raw time-series data are transformed into T1--T4 tasks through event generation or detection, prompt assembly, and format-aware task construction.
}

    \label{fig:pipeline}
\end{figure*}
We introduce \textbf{TemporalBench}, a multi-domain benchmark designed to systematically evaluate agent performance on time-series tasks that require temporal understanding, prediction, and reasoning under contextual and event-driven settings. Rather than treating time-series evaluation as a single forecasting problem, TemporalBench adopts a task-centric design that explicitly probes different temporal competencies through controlled task construction.

TemporalBench is built upon four real-world time-series datasets, spanning retail (\textbf{FreshRetailNet-50K}\cite{wang2025freshretailnet}), healthcare (\textbf{MIMIC-IV}\cite{johnson2023mimic}), energy (\textbf{PSML}\cite{zheng2022multi}), and physical system domains (\textbf{Causal Chambers}\cite{gamella2024causal}). Across these heterogeneous sources, we apply a \emph{unified benchmark construction pipeline} that transforms raw numerical time-series into a structured set of evaluation tasks. An overview of this pipeline is illustrated in Figure~\ref{fig:pipeline}, which shows how event generation or detection, prompt assembly, and format-aware task construction jointly give rise to the four task families (T1–T4).

The resulting benchmark comprises \textbf{2,775} evaluation tasks instantiated over \textbf{191} distinct time-series instances.
For each instance, multiple tasks are generated to probe complementary aspects of temporal intelligence, including intrinsic structure understanding, future prediction, contextual reasoning, and event-informed decision-making.

As summarized in Table~\ref{tab:ts-benchmarks-comparison}, TemporalBench differs from prior temporal and time-series benchmarks by jointly incorporating real numerical time-series signals, explicit contextual information, and reasoning-oriented task formulations across multiple task types.
Tasks are instantiated as either multiple-choice reasoning questions or numerical forecasting objectives, enabling fine-grained analysis of agent behavior across distinct temporal reasoning regimes.

In the following sections, we describe the task taxonomy and benchmark construction pipeline in detail.


\subsection{Task Families}
We organize TemporalBench into four task families (T1--T4) to examine distinct aspects of temporal reasoning under progressively relaxed assumptions, rather than to incrementally increase task difficulty.
The four families form a coherent evaluation sequence that probes how temporal understanding and forecasting behavior change as additional sources of information are introduced.
Concretely, tasks vary along two key factors: whether the focus is on \textbf{interpreting historical behavior} or \textbf{predicting future outcomes}, and whether reasoning is performed using only \textbf{numerical time-series signals} or additionally conditioned on \textbf{contextual and event information}.
For each time-series instance, data are partitioned into historical and future segments using a predefined event boundary, and a complete set of tasks spanning all four families is generated.
T1 and T2 operate solely on numerical time-series signals, while T3 and T4 further incorporate textual context reflecting realistic application scenarios.
This unified construction enables controlled comparison across task families, ensuring that observed performance differences arise from changes in temporal and contextual reasoning requirements rather than from uncontrolled variation in data or task format.

\paragraph{T1: Historical Time-Series Understanding.}
T1 tasks evaluate whether an agent can correctly interpret the structure and behavior of a time series based solely on historical observations.
Agents are provided with a historical time-series segment and asked multiple-choice questions about intrinsic temporal properties, including trend direction, volatility level, seasonality, and the presence of anomalies.
By isolating historical interpretation from prediction and context, T1 serves as a probe of an agent’s ability to extract and reason about temporal structure itself.

\paragraph{T2: Future Time-Series Prediction without Context.}
T2 tasks focus on predicting the future segment of a time series using historical observations alone.
Agents are required to either produce numerical forecasts or answer multiple-choice questions concerning anticipated trends, volatility changes, or seasonal patterns.
By excluding contextual information, T2 aligns with traditional forecasting settings while enabling direct comparison between numerical accuracy and qualitative expectations.
Importantly, T1 and T2 do not form an increasing difficulty hierarchy; rather, they isolate complementary competencies related to interpreting past structure and extrapolating future behavior from temporal signals alone.

\paragraph{T3: Contextual Reasoning over Time-Series.}
T3 tasks introduce a fundamental shift from pattern interpretation to contextual temporal reasoning.
In addition to historical time-series data, agents are provided with textual context that grounds the series in domain semantics and real-world conditions.
Rather than asking what the series looks like, T3 questions probe whether an agent can explain, compare, and reason about temporal behavior in light of contextual descriptions.

To systematically localize reasoning failures, T3 tasks are constructed along six capability dimensions, which together form a minimal basis set for contextual temporal reasoning:
\textbf{C1 Alignment} (mapping natural-language conditions to temporal fields or windows),
\textbf{C2 Slicing} (comparing temporal segments across periods or states),
\textbf{C3 Difference Judgment} (reasoning about distributional differences),
\textbf{C4 Lag} (identifying delayed or offset effects),
\textbf{C5 Structure} (recognizing higher-level temporal patterns such as peaks and change points), and
\textbf{C6 Interaction Understanding} (reasoning about joint or interaction effects).
For each time-series instance, multiple questions are generated to cover a subset of these dimensions, enabling fine-grained attribution of reasoning errors.

\paragraph{T4: Event-Informed Time-Series Prediction.}
T4 tasks evaluate event-conditioned and counterfactual temporal reasoning by requiring agents to predict how future time-series behavior would change under specified upcoming events.
Along with historical observations, agents receive background information describing the application context and textual descriptions of future events expected to influence the series, such as extreme weather or other domain-specific interventions.
Agents must integrate historical patterns, contextual semantics, and event information to produce numerical forecasts or qualitative judgments about future trends, volatility, or seasonality.
T4 therefore probes whether temporal information can be reused for conditional reasoning, rather than merely extrapolated under unchanged assumptions.

\paragraph{Diagnostic Role.}
Together, T1–T4 enable TemporalBench to separate numerical prediction from temporal interpretation, and to further test whether temporal information can be aligned with context and used for reasoning under changing conditions.
This decomposition allows experimental results to be traced back to specific temporal competencies, a property that is essential for diagnosing failure modes in agentic time-series systems.

\subsection{Dataset Transformation Pipeline}

We construct TemporalBench through a unified dataset transformation pipeline designed to ensure methodological consistency, reproducibility, and comparability across heterogeneous time-series domains. The pipeline consists of two key components: (i) event injection or detection to define historical–future boundaries, and (ii) ground truth generation to support systematic evaluation across task families.

\subsubsection{Event Injection and Detection}

Events play a central role in TemporalBench by defining the boundary between historical and future segments of each time series and enabling event-centric task construction.
However, the four datasets in our benchmark differ substantially in how events are recorded or manifested.
Some datasets contain dense temporal measurements with sparse or unreliable event annotations, while others naturally exhibit explicit events or intervention signals.

To accommodate this heterogeneity within a unified benchmark design, we adopt two complementary strategies: \emph{event injection} and \emph{event detection}.
The goal of event injection is not to faithfully simulate real-world causal mechanisms, but to introduce controlled and interpretable regime boundaries that support systematic evaluation of event-aware temporal reasoning.
Accordingly, evaluation focuses on whether agents appropriately condition their predictions and decisions on the presence of an event, rather than on the realism or magnitude of the injected effect.

FreshRetailNet-50K and PSML lack reliable event annotations relative to their dense temporal records.
For these datasets, we inject a salient event at a selected time point and simulate its impact on the subsequent segment.
To reduce bias from imperfect simulation, injected events are designed to be coarse and distinctive, and downstream tasks emphasize qualitative changes between historical and future segments, such as shifts in trend, volatility, or seasonality, rather than precise numerical effects.

In contrast, MIMIC-IV and Causal Chambers naturally support event detection.
MIMIC-IV provides explicit clinical event annotations, including procedures, medication administration, and patient transfers.
Causal Chambers does not include explicit event labels, but interventions in the underlying physical system often induce abrupt changes in observed variables, which we detect as change points.
Under both strategies, each time series is partitioned into a historical segment and a future segment by a well-defined event, enabling consistent task construction across heterogeneous datasets.

\subsubsection{Ground Truth Generation}

Ground truth labels in TemporalBench are generated automatically from time-series data using unified, rule-based procedures, without manual annotation or model-dependent heuristics.
Label generation follows three core principles: independence from contextual descriptions, robustness across heterogeneous domains, and explicit handling of uncertainty.

Across all datasets and task families, labels are computed solely from historical and future time-series segments.
Contextual descriptions and event narratives are used only to define the historical--future split and to condition task prompts, and they do not influence any labeling rule.
This design prevents information leakage from event injection or detection and ensures that models cannot succeed by exploiting prompt structure rather than temporal evidence.

To ensure robustness across domains and scales, we rely on statistically stable estimators, including medians, median absolute deviation (MAD), interquartile range (IQR), and Theil--Sen slopes.
Labels are defined primarily in terms of relative changes rather than absolute values.
When observed effects fail to meet predefined strength or support criteria, we assign \emph{Uncertain} labels instead of forcing unreliable decisions.

Ground truth generation is task-specific.
T1 targets intrinsic properties of historical segments, such as trend, volatility, seasonality, and anomalies, using robust statistics and window-based comparisons.
T2 and T4 focus on future behavior, with numerical ground truth given by observed future values and auxiliary multiple-choice labels describing qualitative changes relative to history.
T3 emphasizes higher-level temporal reasoning over historical data, generating labels for questions involving segment comparisons, lagged responses, and interaction effects, with only statistically reliable instances retained.

Additional implementation details and pseudocode for event injection or detection and ground-truth labeling are provided in Appendix~\ref{app:pseudocode}.

\section{Experiments}
\begin{table*}[t]
\centering
\footnotesize

\begin{subtable}[t]{\textwidth}
\centering
\setlength{\tabcolsep}{2.8pt}
\renewcommand{\arraystretch}{1.2}

\begin{tabular}{c|c|cccc|cccc|cccc|cccc}
\toprule
\multirow{2}{*}{Agent} & Dataset
& \multicolumn{4}{c|}{FreshRetailNet}
& \multicolumn{4}{c|}{PSML}
& \multicolumn{4}{c|}{Causal Chambers}
& \multicolumn{4}{c}{MIMIC} \\
\cmidrule(lr){3-6}
\cmidrule(lr){7-10}
\cmidrule(lr){11-14}
\cmidrule(lr){15-18}

& Task
& T1 & T2 & T3 & \multicolumn{1}{c|}{T4}
& T1 & T2 & T3 & \multicolumn{1}{c|}{T4}
& T1 & T2 & T3 & \multicolumn{1}{c|}{T4}
& T1 & T2 & T3 & T4 \\
\midrule

\multirow{2}{*}{Single LLM}
& SR
& 100\% & 100\% & 84.66\% & \multicolumn{1}{c|}{100\%}
& 100\% & 100\% & 100\% & \multicolumn{1}{c|}{96\%}
& 100\% & 100\% & 100\% & \multicolumn{1}{c|}{100\%}
& 100\% & 100\% & 100\% & 100\% \\
& ACC
& 63.64\% & 52.27\% & 2.89\% & \multicolumn{1}{c|}{13.64\%}
& 67.50\% & 20.67\% & 34.80\% & \multicolumn{1}{c|}{36.00\%}
& 13.33\% & 27.33\% & 35.20\% & \multicolumn{1}{c|}{26.00\%}
& 46.81\% & 21.28\% & 36.61\% & 29.79\% \\
\midrule

\multirow{2}{*}{\makecell[c]{TimeSeries\\Scientist}}
& SR
& 85.80\% & 100.00\% & 99.43\% & \multicolumn{1}{c|}{100.00\%}
& 83.00\% & 100.00\% & 100.00\% & \multicolumn{1}{c|}{100.00\%}
& 100.00\% & 100.00\% & 100.00\% & \multicolumn{1}{c|}{100.00\%}
& 86.17\% & 100.00\% & 100.00\% & 100.00\% \\
& ACC
& 33.52\% & 56.82\% & 3.41\% & \multicolumn{1}{c|}{56.82\%}
& 28.00\% & 26.67\% & 21.60\% & \multicolumn{1}{c|}{27.33\%}
& 28.67\% & 2.67\% & 21.60\% & \multicolumn{1}{c|}{2.67\%}
& 10.11\% & 23.40\% & 28.87\% & 23.40\% \\
\midrule

\multirow{2}{*}{AgentScope}
& SR
& 100.00\% & 100.00\% & 96.59\% & \multicolumn{1}{c|}{100.00\%}
& 100.00\% & 98.00\% & 100.00\% & \multicolumn{1}{c|}{98.00\%}
& 75.00\% & 100.00\% & 99.60\% & \multicolumn{1}{c|}{100.00\%}
& 100.00\% & 100.00\% & 100.00\% & 100.00\% \\
& ACC
& 62.50\% & 12.12\% & 13.64\% & \multicolumn{1}{c|}{18.94\%}
& 66.00\% & 24.67\% & 27.20\% & \multicolumn{1}{c|}{35.33\%}
& 12.00\% & 46.00\% & 44.00\% & \multicolumn{1}{c|}{32.00\%}
& 44.68\% & 21.28\% & 23.95\% & 22.70\% \\
\midrule

\multirow{2}{*}{MetaGPT}
& SR
& 100.00\% & 100.00\% & 94.32\% & \multicolumn{1}{c|}{100.00\%}
& 100.00\% & 98.00\% & 100.00\% & \multicolumn{1}{c|}{100.00\%}
& 75.00\% & 100.00\% & 100.00\% & \multicolumn{1}{c|}{100.00\%}
& 100.00\% & 100.00\% & 100.00\% & 100.00\% \\
& ACC
& 62.50\% & 9.09\% & 5.11\% & \multicolumn{1}{c|}{14.39\%}
& 67.50\% & 21.09\% & 22.00\% & \multicolumn{1}{c|}{31.33\%}
& 10.67\% & 59.33\% & 45.20\% & \multicolumn{1}{c|}{16.00\%}
& 45.74\% & 17.02\% & 28.97\% & 25.53\% \\
\midrule

\multirow{2}{*}{CAMEL}
& SR
& 100.00\% & 100.00\% & 92.05\% & \multicolumn{1}{c|}{100.00\%}
& 100.00\% & 100.00\% & 100.00\% & \multicolumn{1}{c|}{100.00\%}
& 75.00\% & 100.00\% & 100.00\% & \multicolumn{1}{c|}{100.00\%}
& 100.00\% & 100.00\% & 100.00\% & 100.00\% \\
& ACC
& 64.20\% & 0.76\% & 6.25\% & \multicolumn{1}{c|}{31.06\%}
& 68.50\% & 14.00\% & 18.40\% & \multicolumn{1}{c|}{30.67\%}
& 10.00\% & 66.00\% & 42.00\% & \multicolumn{1}{c|}{26.67\%}
& 46.81\% & 20.57\% & 30.14\% & 23.40\% \\
\bottomrule
\end{tabular}

\caption{Multi-choice task performance across datasets.}
\end{subtable}

\vspace{6pt}

\begin{subtable}[t]{\textwidth}
\centering
\setlength{\tabcolsep}{12pt}
\renewcommand{\arraystretch}{1}

\begin{tabular}{c|c|cc|cc|cc!{\vrule width 1.2pt}cc}
\toprule
\multirow{2}{*}{Agent} & \multirow{2}{*}{Metric}
& \multicolumn{2}{c|}{FreshRetailNet}
& \multicolumn{2}{c|}{PSML}
& \multicolumn{2}{c!{\vrule width 1.2pt}}{Causal Chambers}
& \multicolumn{2}{c}{MIMIC} \\
\cmidrule(lr){3-4}
\cmidrule(lr){5-6}
\cmidrule(lr){7-8}
\cmidrule(lr){9-10}

&
& T2 & T4
& T2 & T4
& T2 & T4
& T2 & T4 \\
\midrule

\multirow{3}{*}{Single LLM}
& SR
& 97.73\% & 79.55\%
& 14.00\% & 38.00\%
& 44.00\% & 34.00\%
& 91.49\% & 93.62\% \\
& MAE / OW\_sMAPE
& 0.12 & 0.34
& 0.61 & 0.44
& 2.48 & 2.58
& 15.20 & 16.86 \\
& sMAPE / OW\_RMSSE
& 1.27 & 1.29
& 0.60 & 0.37
& 2.57E{-}05 & 2.69E{-}05
& 0.55 & 0.63 \\
\midrule

\multirow{3}{*}{\makecell[c]{TimeSeries\\Scientist}}
& SR
& 43.18\% & 79.55\%
& 26.00\% & 52.00\%
& 50.00\% & 32.00\%
& 89.36\% & 89.36\% \\
& MAE / OW\_sMAPE
& 0.35 & 0.51
& 1.53 & 0.84
& 2.44 & 2.94
& 15.81 & 17.18 \\
& sMAPE / OW\_RMSSE
& 1.27 & 1.40
& 0.65 & 0.48
& 2.53E{-}05 & 3.06E{-}05
& 0.52 & 0.64 \\
\midrule

\multirow{3}{*}{AgentScope}
& SR
& 41.94\% & 23.94\%
& 5.26\% & 31.58\%
& 45.10\% & 27.45\%
& 93.75\% & 89.80\% \\
& MAE / OW\_sMAPE
& 0.12 & 0.20
& 0.28 & 0.35
& 2.76 & 2.66
& 11.05 & 12.02 \\
& sMAPE / OW\_RMSSE
& 126.27 & 130.86
& 37.38 & 30.51
& 2.62E{-}03 & 2.46E{-}03
& 0.43 & 0.49 \\
\midrule

\multirow{3}{*}{MetaGPT}
& SR
& 90.91\% & 88.64\%
& 16.00\% & 64.00\%
& 52.00\% & 40.00\%
& 93.62\% & 93.62\% \\
& MAE / OW\_sMAPE
& 0.13 & 0.24
& 0.34 & 0.40
& 2.62 & 2.76
& 14.11 & 15.40 \\
& sMAPE / OW\_RMSSE
& 126.59 & 127.22
& 24.74 & 43.47
& 2.72E{-}03 & 2.87E{-}03
& 0.53 & 0.63 \\
\midrule

\multirow{3}{*}{CAMEL}
& SR
& 93.18\% & 93.18\%
& 34.00\% & 62.00\%
& 52.00\% & 30.00\%
& 95.74\% & 93.62\% \\
& MAE / OW\_sMAPE
& 0.13 & 0.28
& 0.43 & 0.45
& 2.99 & 2.50
& 12.02 & 15.74 \\
& sMAPE / OW\_RMSSE
& 126.75 & 128.18
& 34.89 & 35.78
& 3.11E{-}03 & 2.60E{-}03
& 0.55 & 0.59 \\
\bottomrule
\end{tabular}

\caption{Forecasting performance on T2 and T4 tasks across datasets.}
\end{subtable}

\caption{Performance of different agent frameworks across multi-choice and forecasting tasks using \texttt{gpt-4o} as the base model.}
\label{tab:main_results}
\vspace{-20pt}
\end{table*}
The experiments in this section are designed to systematically evaluate the temporal competencies isolated by the four task families in TemporalBench. Rather than treating performance as a single aggregate score, we analyze agent behavior across T1–T4 to disentangle distinct aspects of temporal intelligence.

By evaluating different agent paradigms under identical inputs and constraints, our experiments aim to answer three guiding questions: (i) whether strong numerical forecasting performance translates into reliable temporal interpretation, (ii) how the introduction of contextual information alters reasoning behavior over time-series data, and (iii) whether agents can meaningfully adapt their predictions under explicitly specified future events. Each experiment is therefore diagnostic in nature, with observed failure modes mapped back to specific task families and reasoning dimensions rather than treated as isolated errors.

\subsection{Experiment Setup}

Our experimental design aims to evaluate agentic temporal reasoning capabilities in a controlled and diagnostic manner, rather than to optimize predictive accuracy. In particular, the setup is designed to disentangle the effects of agent architecture, backbone language model capacity, and evaluation protocol across different task families in TemporalBench.

\subsubsection{Models and Agents}

We evaluate both time-series--specialized agents and general-purpose LLM-based agents to assess their performance on the proposed benchmark.
Given the limited availability of agent systems explicitly designed for time-series tasks, we include \textbf{TimeSeriesScientist}~\cite{zhao2025timeseriesscientist} as a representative time-series--specific agent.
We emphasize that TimeSeriesScientist is used to represent domain-specialized, structure-aware agentic approaches, rather than as an exhaustive benchmark of time-series modeling techniques.

In addition, we evaluate several general-purpose agent frameworks that have been widely adopted for reasoning and decision-making tasks, including \textbf{MetaGPT}~\cite{hong2024metagpt}, \textbf{AgentScope}~\cite{gao2024agentscope}, and \textbf{CAMEL}~\cite{li2023camel}.
These frameworks are not specialized for time-series data and primarily rely on language-based reasoning and tool orchestration, making them suitable baselines for assessing temporal reasoning in the absence of domain-specific inductive biases.

To isolate the contribution of agent frameworks themselves, we further include a \emph{direct LLM prompting} baseline, in which a single large language model is queried without agent structure, tool use, or multi-step interaction.
This setting reflects common practice and provides a reference point for comparing agent-based approaches against raw LLM capabilities.

All agents and baselines are instantiated with a common set of backbone language models to ensure controlled comparison.
The backbone models include \textbf{GPT-4o}~\cite{hurst2024gpt}, \textbf{Gemini-2.5-Flash}~\cite{gemini25flash}, \textbf{Claude-3-7-Sonnet}~\cite{claude37sonnet}, \textbf{DeepSeek-Chat}~\cite{liu2024deepseek}, and \textbf{Qwen-Plus}~\cite{bai2023qwen}.
Across all settings, agents receive identical task inputs, output constraints, and access to information, ensuring that performance differences reflect agent design rather than model capacity.

We do not treat standalone forecasting models as primary baselines, as TemporalBench targets agentic temporal reasoning and decision-making rather than pure predictive accuracy.
Traditional forecasting performance is therefore used as a reference signal rather than a proxy for task competence.

\subsubsection{Evaluation Metrics}

We adopt evaluation metrics that reflect the distinct objectives of reasoning-oriented and prediction-oriented tasks in TemporalBench. For multiple-choice question-answering tasks, including T1, T3, and the qualitative components of T2 and T4, we report \emph{accuracy} as the primary metric. Accuracy is chosen to emphasize unambiguous decision correctness, aligning with the benchmark’s goal of identifying discrete reasoning failures rather than grading partial or stylistic explanations.

Questions labeled as \emph{Uncertain} or \emph{Inconclusive} are excluded from accuracy computation for all models. This ensures that evaluation focuses on instances with sufficient statistical support and avoids penalizing models on inherently ambiguous cases.

For numerical forecasting tasks in T2 and T4, we evaluate prediction quality using error-based metrics computed between predicted and ground-truth future values over the valid forecast horizon. For datasets involving single or low-dimensional target series (e.g., \textbf{FreshRetailNet}, \textbf{PSML}, and \textbf{Causal Chambers}), we report standard forecasting metrics including mean absolute error (MAE) and symmetric mean absolute percentage error (sMAPE).

For the \textbf{MIMIC} dataset, forecasting involves multiple correlated clinical time series with heterogeneous scales.
To ensure fair aggregation across series and avoid domination by high-variance signals, we use \emph{overall weighted} metrics, specifically OW\_sMAPE and OW\_RMSSE.
These metrics are standard in multi-series forecasting and enable fair comparison across differing scales and prediction horizons.

\subsection{Main Results}

Table~\ref{tab:main_results} summarizes the performance of different agent frameworks and a single-LLM baseline across four task families (T1--T4) and four domains, using \texttt{gpt-4o} as the backbone model. 

\subsubsection{Global Performance Patterns Across Task Families}
Here we analyze how performance evolves across task families as temporal reasoning requirements shift from structural interpretation (T1), to pure prediction without context (T2), to contextual and event-aware reasoning (T3 and T4). Across all evaluated methods, TemporalBench reveals a consistent stratification aligned with this design.

Historical understanding tasks (T1) achieve moderate accuracy for most agents, indicating that intrinsic temporal properties such as trend direction or volatility are partially accessible when reasoning is confined to historical structure alone. In contrast, reasoning-intensive tasks that require grounding temporal patterns in context (T3) remain uniformly challenging: accuracy frequently drops to single-digit or low double-digit percentages in retail and clinical domains, even when the same models perform reasonably well on T1. This sharp degradation highlights the difficulty of contextual temporal reasoning beyond structural interpretation.

Prediction-related multiple-choice tasks (T2 and T4) exhibit substantially higher variance across agents than historical understanding tasks. Notably, inferring future behavior in a discrete, qualitative label space proves markedly more difficult than producing numerical forecasts. Across datasets, improvements in numerical forecasting accuracy (e.g., lower MAE or sMAPE) do not reliably translate into higher accuracy on qualitative future-oriented judgments. This decoupling between numerical prediction quality and qualitative temporal interpretation directly reflects the distinction that TemporalBench is designed to expose.

\begin{keydiscovery1}
\textbf{Key Discovery.}
Temporal reasoning performance degrades sharply as tasks move from structural understanding (T1) to contextual reasoning (T3), and strong numerical forecasting accuracy does not reliably translate into accurate qualitative judgments about future behavior (T2/T4).
\end{keydiscovery1}

\subsubsection{Agent Frameworks versus Direct LLM Prompting}
This subsection evaluates whether agent orchestration provides systematic benefits over direct LLM prompting across the temporal competencies isolated by T1--T4. Comparing agent-based methods with direct LLM prompting reveals that agent frameworks do not consistently improve temporal reasoning performance.

On qualitative tasks that require either historical interpretation (T1) or context-free future reasoning (T2), the single-LLM baseline frequently matches or exceeds the accuracy of general-purpose agents, particularly on FreshRetailNet and MIMIC. These results suggest that when temporal structure must be inferred directly from numerical sequences, additional agent steps can introduce reasoning noise rather than providing effective decomposition or abstraction.

Forecasting tasks present a more nuanced pattern. While the single LLM achieves competitive MAE and sMAPE on relatively structured domains such as FreshRetailNet and MIMIC, agent-based methods occasionally reduce numerical error on PSML and Causal Chambers. However, these gains are inconsistent across domains and task families, indicating that current agent frameworks lack a robust and general mechanism for exploiting temporal regularities beyond what is already captured by the backbone model.

\begin{keydiscovery1}
\textbf{Key Discovery.}
Agent orchestration alone does not guarantee improved temporal reasoning, and can introduce additional noise when temporal structure must be inferred directly from numerical time-series data.
\end{keydiscovery1}

\subsubsection{Time-Series-Specific Agent versus General-Purpose Agents}
This subsection contrasts domain-specific temporal modeling with general-purpose agent reasoning across the disentangled competencies evaluated by TemporalBench. The time-series--specific agent does not consistently outperform general-purpose agents across task families, but it exhibits a distinct and systematic performance profile.

On prediction-oriented tasks (T2 and T4), its error metrics are comparatively stable across domains, remaining within a narrower range than those of general-purpose agents. This stability aligns with its design emphasis on explicit time-series modeling and numerical consistency. However, this advantage does not extend to qualitative reasoning tasks.

On multiple-choice tasks, particularly those requiring contextual grounding and higher-level reasoning (T3), the time-series-specific agent offers limited improvement and often underperforms general-purpose agents. These results indicate that while specialized temporal architectures can stabilize numerical prediction, they do not directly address challenges such as abstraction, comparison across temporal segments, or conditional reasoning under context, which are central to T3 and T4.

\begin{keydiscovery1}
\textbf{Key Discovery.}
Domain-specific temporal modeling stabilizes numerical prediction (T2/T4) but does not address higher-level reasoning challenges such as abstraction, comparison, or conditional reasoning under context (T3).
\end{keydiscovery1}

\subsubsection{Domain-Specific Effects and Event-Aware Reasoning}
This subsection examines how domain characteristics interact with the event-aware reasoning requirements introduced in T4. Performance varies substantially across domains. Physical and semi-synthetic datasets such as PSML and Causal Chambers generally yield higher accuracy and more stable forecasting errors, reflecting clearer temporal structure and stronger signal-to-noise ratios. In contrast, FreshRetailNet and MIMIC exhibit consistently lower accuracy, particularly on reasoning-intensive (T3) and event-aware (T4) tasks, highlighting the challenges posed by noisy dynamics, confounding factors, and heterogeneous event effects.

Event-informed prediction tasks further expose limitations in how agents utilize contextual and event information. While incorporating event descriptions occasionally improves numerical forecasting relative to context-free prediction (T2) in specific domains, qualitative accuracy does not consistently benefit from the presence of event narratives. This pattern suggests that current agents struggle to translate event descriptions into actionable temporal constraints, underscoring the importance of explicitly evaluating event-conditioned temporal reasoning rather than assuming that contextual integration emerges implicitly.

\begin{keydiscovery1}
\textbf{Key Discovery.}
Event descriptions are not consistently translated into actionable temporal constraints by current agents, revealing a gap between contextual awareness and event-conditioned temporal reasoning (T4).
\end{keydiscovery1}

Overall, these results show that existing agent frameworks exhibit fragmented strengths across the temporal competencies evaluated by T1--T4. By disentangling historical understanding, prediction, contextual reasoning, and event-conditioned forecasting, TemporalBench reveals gaps between numerical accuracy, temporal interpretation, and event-aware reasoning that are largely hidden by forecasting-only benchmarks.


\section{Analysis}

Although accuracy on some reasoning-oriented tasks is low, TemporalBench does not exhibit a uniform floor effect. Instead, performance varies systematically across task families, domains, and agent designs, revealing structured capability differences. In this section, we analyze these patterns to characterize how temporal competencies emerge and fail under controlled settings.

\subsection{Analysis of Temporal Reasoning}

\begin{figure}[htbp]
\vspace{-10pt}
    \centering
 \includegraphics[width=\columnwidth]{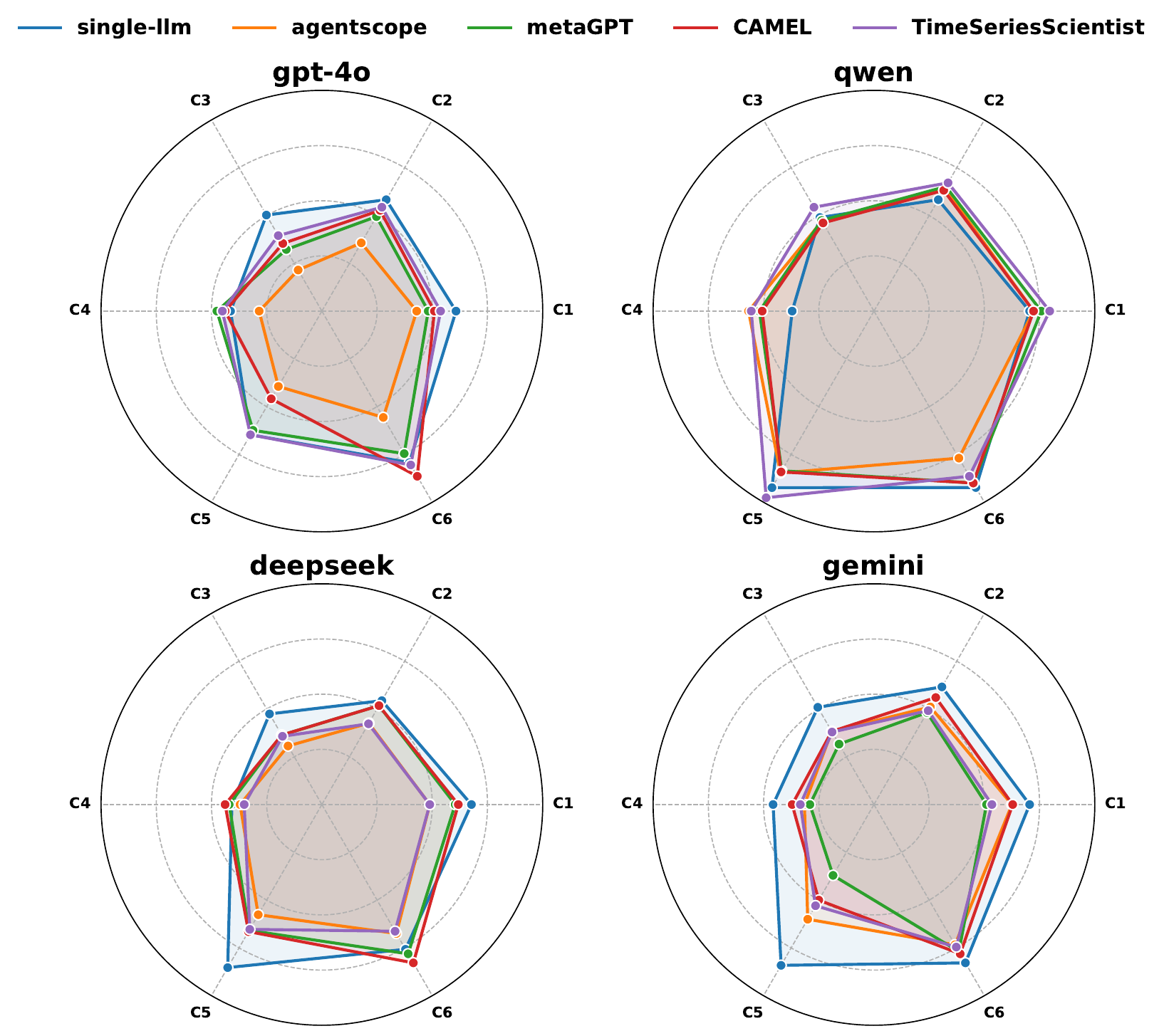}
    \caption{
Radar plots showing the performance of different agents on the six T3 reasoning dimensions (C1--C6) under different base LLMs.
Each subplot corresponds to a base model, and each curve represents an agent.
}

    \label{fig:t3_radar}
    \vspace{-10pt}
\end{figure}
To further understand agent reasoning behavior beyond aggregate accuracy, we analyze performance on T3 tasks by decomposing them into six reasoning capabilities (C1–C6). Figure~\ref{fig:t3_radar} visualizes accuracy across all agents and four backbone models. Rather than exhibiting uniform improvements, agents show highly uneven capability profiles.

Across all backbone models, C6 (interaction understanding) and C5 (structural pattern recognition) consistently achieve higher accuracy than other capabilities. These tasks primarily rely on recognizing co-occurring conditions, peak patterns, or regime-level changes, which are more amenable to surface-level pattern matching and statistical cues. In contrast, C2 (slicing-based comparison) and C3 (difference judgment) remain challenging for most agents, indicating persistent difficulty in reliably comparing temporal segments or quantifying relative changes.

Comparing agent types, general-purpose agents such as MetaGPT and CAMEL exhibit modest gains over the single-LLM baseline on higher-level reasoning dimensions (e.g., C5 and C6), but often underperform on lower-level analytical capabilities such as C1 (alignment) and C4 (lag detection). This suggests that language-driven orchestration can help with abstract interpretation but does not reliably improve fine-grained temporal analysis.

The time-series–specific agent shows a distinct but limited advantage profile. While its performance does not dominate across all six capabilities, it demonstrates more balanced behavior on C1 and C4, which require explicit temporal alignment and response-window reasoning. This pattern is consistent with its design emphasis on structured time-series processing. However, it does not substantially improve performance on higher-level comparative or interaction-based reasoning, reinforcing that specialized temporal modeling alone is insufficient for complex reasoning tasks.

Overall, the C1–C6 analysis reveals that current agents possess fragmented reasoning capabilities rather than unified temporal intelligence. Strengths in interaction and structure recognition coexist with weaknesses in comparative and causal reasoning, highlighting the need for benchmarks that disentangle reasoning dimensions and for agent designs that explicitly address these gaps.

\begin{insightbox}
\textbf{Insight.}
Temporal reasoning in current agents appears non-uniform, with structural and interaction reasoning (C5–C6) more accessible than comparative and distributional reasoning (C2–C3), suggesting that T3 failures are linked to specific reasoning gaps rather than data scarcity or model size.
\end{insightbox}

\subsection{The Role of Temporal Representations in Agent Reasoning}

To further diagnose the limitations of general-purpose agents on temporal tasks, we design a lightweight feature extraction and visualization module and integrate it into a representative general agent, AgentScope with \texttt{gpt-4o} as the backbone.
In contrast to time-series–specific agents, general agents typically ingest temporal data by directly serializing numerical values into long textual prompts, without mechanisms to explicitly expose temporal structure.
Such raw serialization obscures relationships among values, making it difficult for LLMs to reason about trends, volatility changes, or structural shifts over time.

\paragraph{Module Design}
The proposed module augments raw time-series input with two complementary representations.
First, we extract a compact set of structured numerical features summarizing global statistics and multi-scale temporal dynamics.
This includes a reduced subset of interpretable catch22 features~\cite{lubba2019catch22} capturing distributional properties, complexity, and outliers, as well as multi-scale structural features such as short-, mid-, and long-term slopes, recent volatility, and shift indicators.
All features are described in natural language within the prompt to ensure interpretability.
Second, we generate a visualization of the time series as an additional modality, exposing relative changes, peaks, and structural patterns that are difficult to infer from raw numeric tokens alone.

\paragraph{Impact on Temporal Understanding}
We evaluate the effect of this module on AgentScope by comparing performance under different input representations.
Table~\ref{tab:feature_visual_ablation} reports results for raw time-series inputs, feature-based augmentation, visualization-based augmentation, and their combination.
Overall, the impact of enhanced representations is heterogeneous across domains and task families rather than uniformly positive.

In domains with clearer temporal structure and stronger signal-to-noise ratios, such as \emph{Causal Chambers}, enhanced representations lead to consistent improvements on multi-choice prediction and reasoning tasks, particularly for future-related judgments (T2) and event-aware prediction (T4).
In contrast, for more complex and noisy real-world domains such as \emph{FreshRetailNet} and \emph{PSML}, gains are limited and sometimes mixed:
while historical understanding (T1) often benefits from explicit structural cues, performance on reasoning-intensive tasks (T3) and future multi-choice prediction (T2) does not consistently improve and may even degrade in some settings.

These results suggest that explicitly exposing temporal structure through features and visualizations can alleviate certain failure modes of general agents, especially in structured domains.
However, representation-level augmentation alone is insufficient to resolve broader challenges of temporal reasoning and decision-making in complex real-world time-series.
Importantly, all observed effects are obtained without additional training, task-specific tuning, or changes to the agent workflow, indicating that representation quality is a meaningful but partial factor in agent performance on time-series tasks.

\begin{insightbox}
\textbf{Insight.}
Explicit temporal representations can partially alleviate reasoning failures, but their effectiveness is domain- and task-dependent.
Improvements emerge primarily in settings with clearer temporal structure, suggesting that representation quality is a necessary but insufficient condition for robust contextual and event-aware temporal reasoning.
\end{insightbox}

\begin{table}[htbp]
\vspace{-10pt}
\centering
\footnotesize
\setlength{\tabcolsep}{2.6pt}
\renewcommand{\arraystretch}{1}

\newcommand{\rowvbar}{\raisebox{-0.55ex}{\rule{0.45pt}{2.6ex}}}

\begin{tabular}{lcccc c cccc}
\toprule
& \multicolumn{4}{c}{\textbf{FreshRetailNet}} & & \multicolumn{4}{c}{\textbf{PSML}} \\
\cmidrule(lr){2-5}\cmidrule(lr){7-10}
\textbf{} & \textbf{T1} & \textbf{T2} & \textbf{T3} & \textbf{T4} & \rowvbar
& \textbf{T1} & \textbf{T2} & \textbf{T3} & \textbf{T4} \\
\midrule
\textbf{Raw}      & 62.50\% & 12.12\% & \textbf{13.64\%} & 18.94\% & \rowvbar & 66.00\% & 24.67\% & \textbf{27.20\%} & 35.33\% \\
\textbf{Feat.}       & 64.20\% & \textbf{16.67\%} & \textbf{13.64\%} & 18.94\% & \rowvbar & 63.00\% & \textbf{30.61\%} & 26.80\% & 29.33\% \\
\textbf{Vis.}   & 63.64\% & 9.09\%  & 10.80\% & \textbf{20.45\%} & \rowvbar & \textbf{67.50\%} & 26.67\% & 22.00\% & 34.67\% \\
 \textbf{Both}   & \textbf{69.32\%} & 5.30\% & 10.80\% & 15.15\% & \rowvbar & 65.00\% & 21.77\% & 21.20\% & \textbf{40.67\%} \\
\midrule
& \multicolumn{4}{c}{\textbf{MIMIC}} & & \multicolumn{4}{c}{\textbf{Causal Chambers}} \\
\cmidrule(lr){2-5}\cmidrule(lr){7-10}
\textbf{} & \textbf{T1} & \textbf{T2} & \textbf{T3} & \textbf{T4} & \rowvbar
& \textbf{T1} & \textbf{T2} & \textbf{T3} & \textbf{T4} \\
\midrule
\textbf{Raw}      & \textbf{44.68\%} & 21.28\% & 23.95\% & 22.70\% & \rowvbar & 12.00\% & 46.00\% & \textbf{44.00\%} & 32.00\% \\
\textbf{Feat.}        & 43.62\% & \textbf{25.53\%} & 24.46\% & 21.99\% & \rowvbar & \textbf{13.33\%} & 36.67\% & 42.40\% & 30.00\% \\
 \textbf{Vis.}   & \textbf{44.68\%} & 22.70\% & \textbf{26.02\%} & 24.11\% & \rowvbar & \textbf{13.33\%} & 41.33\% & 40.00\% & 36.00\% \\
 \textbf{Both}   & 44.15\% & 24.11\% & 25.43\% & \textbf{24.82\%} & \rowvbar & 10.67\% & \textbf{52.67\%} & 42.00\% & \textbf{38.67\%} \\
\bottomrule
\end{tabular}

\caption{Multi-choice accuracy (ACC) of AgentScope with \texttt{gpt-4o} under different temporal representations.
\textit{Feat.} denotes feature-based augmentation, \textit{Vis.} denotes visualization-based augmentation, and \textit{Both} indicates the combination of feature and visualization augmentations.
Bold indicates the best performance within each dataset and task column.}
\label{tab:feature_visual_ablation}

\end{table}
\vspace{-25pt}

\section{Conclusion}

This work introduces \textbf{TemporalBench}, a benchmark for systematically evaluating agent and LLM performance on time-series tasks that require temporal understanding, prediction, and contextual awareness.
By combining real-world time-series data with event-centric task construction and a four-tier task taxonomy (T1--T4), TemporalBench moves beyond traditional forecasting-only evaluations and enables structured assessment of temporal competence under diverse informational settings.

Our empirical results show that strong numerical forecasting performance does not reliably translate into correct decisions when tasks require contextual grounding or event-conditioned reasoning.
Performance varies substantially across task families and domains, indicating that temporal competence is not a single unified capability.
Improvements in forecasting accuracy alone are therefore insufficient to ensure robust behavior in settings that involve contextual interpretation or conditional prediction.

Taken together, these findings suggest that current agentic approaches to time-series problems exhibit fragmented strengths rather than holistic temporal understanding.
While some systems perform well on isolated aspects of temporal tasks, none consistently integrate temporal structure, contextual information, and decision requirements across all evaluated settings.

From a broader perspective, TemporalBench highlights the need for evaluation frameworks that disentangle complementary temporal competencies instead of collapsing them into a single end-to-end objective.
By evaluating agents across multiple task types, domains, and informational regimes, the benchmark provides a foundation for more principled analysis of agent capabilities and limitations in real-world time-series applications.

\bibliographystyle{ACM-Reference-Format}
\bibliography{ref}

@article{qiu2024tfb,
  title={TFB: Towards Comprehensive and Fair Benchmarking of Time Series Forecasting Methods},
  author={Qiu, Xiangfei and Hu, Jilin and Zhou, Lekui and Wu, Xingjian and Du, Junyang and Zhang, Buang and Guo, Chenjuan and Zhou, Aoying and Jensen, Christian S and Sheng, Zhenli and Yang, Bin},
  journal={Proceedings of the VLDB Endowment},
  volume={17},
  number={9},
  pages={2363--2377},
  year={2024}
}

@article{aksu2024gifteval,
  title={GIFT Eval: A Benchmark for General Time Series Forecasting Model Evaluation},
  author={Aksu, Taha and Woo, Gerald and Liu, Juncheng and Liu, Xu and Liu, Chenghao and Savarese, Silvio and Xiong, Caiming and Sahoo, Doyen},
  journal={arXiv preprint arXiv:2409.16086},
  year={2024}
}

@article{ansari2024chronos,
  title={Chronos: Learning the Language of Time Series},
  author={Ansari, Abdul Fatir and T{\"u}rkmen, Caner and Stella, Lorenzo and Shchur, Oleksandr and Turkmen, Funda and others},
  journal={arXiv preprint arXiv:2403.07815},
  year={2024}
}

@inproceedings{chu2024timebench,
  title={TimeBench: A Comprehensive Evaluation of Temporal Reasoning Abilities in Large Language Models},
  author={Chu, Zhenrui and Xu, Yilun and Li, Yikang and Zhang, Hongming and Zhang, Yue},
  booktitle={Proceedings of ACL},
  year={2024}
}

@article{zhao2025timeseriesscientist,
  title={TimeSeriesScientist: A General Purpose AI Agent for Time Series Analysis},
  author={Zhao, Haokun and Zhang, Xiang and Wei, Jiaqi and Xu, Yiwei and He, Yuting and Sun, Siqi and You, Chenyu},
  journal={arXiv preprint arXiv:2510.01538},
  year={2025}
}

@article{jin2024whatcanllm,
  title={What Can Large Language Models Tell Us about Time Series Forecasting?},
  author={Jin, Mingxuan and Zhang, Haixu and Wang, Wenjie and Wang, Yasha and others},
  journal={arXiv preprint arXiv:2411.09089},
  year={2024}
}

@article{hong2024metagpt,
  title={MetaGPT: Meta Programming for a Multi Agent Collaborative Framework},
  author={Hong, Sirui and others},
  journal={arXiv preprint arXiv:2308.00352},
  year={2024}
}

@article{gao2024agentscope,
  title={AgentScope: A Flexible yet Robust Multi Agent Platform},
  author={Gao, Dawei and others},
  journal={arXiv preprint arXiv:2402.14034},
  year={2024}
}

@inproceedings{li2023camel,
  title={CAMEL: Communicative Agents for Mind Exploration of AI Society},
  author={Li, Guohao and others},
  booktitle={NeurIPS},
  year={2023}
}

@article{plaat2025agentic,
  title={Agentic large language models, a survey},
  author={Plaat, Aske and van Duijn, Max and van Stein, Niki and Preuss, Mike and van der Putten, Peter and Batenburg, Kees Joost},
  journal={arXiv preprint arXiv:2503.23037},
  year={2025}
}

@article{jimenez2023swe,
  title={Swe-bench: Can language models resolve real-world github issues?},
  author={Jimenez, Carlos E and Yang, John and Wettig, Alexander and Yao, Shunyu and Pei, Kexin and Press, Ofir and Narasimhan, Karthik},
  journal={arXiv preprint arXiv:2310.06770},
  year={2023}
}

@article{boiko2023emergent,
  title={Emergent autonomous scientific research capabilities of large language models},
  author={Boiko, Daniil A and MacKnight, Robert and Gomes, Gabe},
  journal={arXiv preprint arXiv:2304.05332},
  year={2023}
}

@article{singh2024malmm,
  title={Malmm: Multi-agent large language models for zero-shot robotics manipulation},
  author={Singh, Harsh and Das, Rocktim Jyoti and Han, Mingfei and Nakov, Preslav and Laptev, Ivan},
  journal={arXiv preprint arXiv:2411.17636},
  year={2024}
}

@article{zhang2024large,
  title={Large language models for time series: A survey},
  author={Zhang, Xiyuan and Chowdhury, Ranak Roy and Gupta, Rajesh K and Shang, Jingbo},
  journal={arXiv preprint arXiv:2402.01801},
  year={2024}
}

@article{yeh2025empowering,
  title={Empowering Time Series Forecasting with LLM-Agents},
  author={Yeh, Chin-Chia Michael and Lai, Vivian and Saini, Uday Singh and Fan, Xiran and Fan, Yujie and Wang, Junpeng and Dai, Xin and Zheng, Yan},
  journal={arXiv preprint arXiv:2508.04231},
  year={2025}
}

@article{cao2025timedit,
  title={TimeDiT: Diffusion Transformers Foundation Model for Time Series Forecasting},
  author={Cao, Defu and Liu, Yan},
  year={2025}
}

@article{yang2025foundation,
  title={Foundation Models for Demand Forecasting via Dual-Strategy Ensembling},
  author={Yang, Wei and Cao, Defu and Liu, Yan},
  journal={arXiv preprint arXiv:2507.22053},
  year={2025}
}

@article{ye2024domain,
  title={Domain-Oriented Time Series Inference Agents for Reasoning and Automated Analysis},
  author={Ye, Wen and Yang, Wei and Cao, Defu and Zhang, Yizhou and Tang, Lumingyuan and Cai, Jie and Liu, Yan},
  journal={arXiv preprint arXiv:2410.04047},
  year={2024}
}

@article{cao2025conversational,
  title={Conversational Time Series Foundation Models: Towards Explainable and Effective Forecasting},
  author={Cao, Defu and Gee, Michael and Liu, Jinbo and Wang, Hengxuan and Yang, Wei and Wang, Rui and Liu, Yan},
  journal={arXiv preprint arXiv:2512.16022},
  year={2025}
}

@article{ye2025llm,
  title={When LLM Meets Time Series: Can LLMs Perform Multi-Step Time Series Reasoning and Inference},
  author={Ye, Wen and Liu, Jinbo and Cao, Defu and Yang, Wei and Liu, Yan},
  journal={arXiv preprint arXiv:2509.01822},
  year={2025}
}

@inproceedings{chen2025tourrank,
  title={Tourrank: Utilizing large language models for documents ranking with a tournament-inspired strategy},
  author={Chen, Yiqun and Liu, Qi and Zhang, Yi and Sun, Weiwei and Ma, Xinyu and Yang, Wei and Shi, Daiting and Mao, Jiaxin and Yin, Dawei},
  booktitle={Proceedings of the ACM on Web Conference 2025},
  pages={1638--1652},
  year={2025}
}

@article{ping2025hdlcore,
  title={Hdlcore: A training-free framework for mitigating hallucinations in llm-generated hdl},
  author={Ping, Heng and Li, Shixuan and Zhang, Peiyu and Cheng, Anzhe and Duan, Shukai and Kanakaris, Nikos and Xiao, Xiongye and Yang, Wei and Nazarian, Shahin and Irimia, Andrei and others},
  journal={arXiv preprint arXiv:2503.16528},
  year={2025}
}

@article{ping2025verimoa,
  title={VeriMoA: A Mixture-of-Agents Framework for Spec-to-HDL Generation},
  author={Ping, Heng and Bhattacharjee, Arijit and Zhang, Peiyu and Li, Shixuan and Yang, Wei and Cheng, Anzhe and Zhang, Xiaole and Thomason, Jesse and Jannesari, Ali and Ahmed, Nesreen and others},
  journal={arXiv preprint arXiv:2510.27617},
  year={2025}
}

@article{yang2025maestro,
  title={Maestro: Learning to Collaborate via Conditional Listwise Policy Optimization for Multi-Agent LLMs},
  author={Yang, Wei and Pang, Jiacheng and Li, Shixuan and Bogdan, Paul and Tu, Stephen and Thomason, Jesse},
  journal={arXiv preprint arXiv:2511.06134},
  year={2025}
}

@article{yang2025learning,
  title={Learning to deliberate: Meta-policy collaboration for agentic llms with multi-agent reinforcement learning},
  author={Yang, Wei and Thomason, Jesse},
  journal={arXiv preprint arXiv:2509.03817},
  year={2025}
}

@article{cao2023tempo,
  title={Tempo: Prompt-based generative pre-trained transformer for time series forecasting},
  author={Cao, Defu and Jia, Furong and Arik, Sercan O and Pfister, Tomas and Zheng, Yixiang and Ye, Wen and Liu, Yan},
  journal={arXiv preprint arXiv:2310.04948},
  year={2023}
}

@article{williams2024context,
  title={Context is key: A benchmark for forecasting with essential textual information},
  author={Williams, Andrew Robert and Ashok, Arjun and Marcotte, {\'E}tienne and Zantedeschi, Valentina and Subramanian, Jithendaraa and Riachi, Roland and Requeima, James and Lacoste, Alexandre and Rish, Irina and Chapados, Nicolas and others},
  journal={arXiv preprint arXiv:2410.18959},
  year={2024}
}

@article{wang2024news,
  title={From news to forecast: Integrating event analysis in llm-based time series forecasting with reflection},
  author={Wang, Xinlei and Feng, Maike and Qiu, Jing and Gu, Jinjin and Zhao, Junhua},
  journal={Advances in Neural Information Processing Systems},
  volume={37},
  pages={58118--58153},
  year={2024}
}

@article{ning2020torque,
  title={TORQUE: A reading comprehension dataset of temporal ordering questions},
  author={Ning, Qiang and Wu, Hao and Han, Rujun and Peng, Nanyun and Gardner, Matt and Roth, Dan},
  journal={arXiv preprint arXiv:2005.00242},
  year={2020}
}

@article{fatemi2024test,
  title={Test of time: A benchmark for evaluating llms on temporal reasoning},
  author={Fatemi, Bahare and Kazemi, Mehran and Tsitsulin, Anton and Malkan, Karishma and Yim, Jinyeong and Palowitch, John and Seo, Sungyong and Halcrow, Jonathan and Perozzi, Bryan},
  journal={arXiv preprint arXiv:2406.09170},
  year={2024}
}

@inproceedings{wang2024tram,
  title={Tram: Benchmarking temporal reasoning for large language models},
  author={Wang, Yuqing and Zhao, Yun},
  booktitle={Findings of the Association for Computational Linguistics: ACL 2024},
  pages={6389--6415},
  year={2024}
}

@article{wang2025freshretailnet,
  title={FreshRetailNet-50K: A Stockout-Annotated Censored Demand Dataset for Latent Demand Recovery and Forecasting in Fresh Retail},
  author={Wang, Yangyang and Gu, Jiawei and Long, Li and Li, Xin and Shen, Li and Fu, Zhouyu and Zhou, Xiangjun and Jiang, Xu},
  journal={arXiv preprint arXiv:2505.16319},
  year={2025}
}

@article{johnson2023mimic,
  title={MIMIC-IV, a freely accessible electronic health record dataset},
  author={Johnson, Alistair EW and Bulgarelli, Lucas and Shen, Lu and Gayles, Alvin and Shammout, Ayad and Horng, Steven and Pollard, Tom J and Hao, Sicheng and Moody, Benjamin and Gow, Brian and others},
  journal={Scientific data},
  volume={10},
  number={1},
  pages={1},
  year={2023},
  publisher={Nature Publishing Group UK London}
}

@article{zheng2022multi,
  title={A multi-scale time-series dataset with benchmark for machine learning in decarbonized energy grids},
  author={Zheng, Xiangtian and Xu, Nan and Trinh, Loc and Wu, Dongqi and Huang, Tong and Sivaranjani, S and Liu, Yan and Xie, Le},
  journal={Scientific Data},
  volume={9},
  number={1},
  pages={359},
  year={2022},
  publisher={Nature Publishing Group UK London}
}

@article{gamella2024causal,
  title={The causal chambers: Real physical systems as a testbed for ai methodology},
  author={Gamella, Juan L and Peters, Jonas and B{\"u}hlmann, Peter},
  journal={arXiv preprint arXiv:2404.11341},
  year={2024}
}

@article{du2024tsi,
  title={Tsi-bench: Benchmarking time series imputation},
  author={Du, Wenjie and Wang, Jun and Qian, Linglong and Yang, Yiyuan and Ibrahim, Zina and Liu, Fanxing and Wang, Zepu and Liu, Haoxin and Zhao, Zhiyuan and Zhou, Yingjie and others},
  journal={arXiv preprint arXiv:2406.12747},
  year={2024}
}

@article{liu2024elephant,
  title={The elephant in the room: Towards a reliable time-series anomaly detection benchmark},
  author={Liu, Qinghua and Paparrizos, John},
  journal={Advances in Neural Information Processing Systems},
  volume={37},
  pages={108231--108261},
  year={2024}
}

@article{liu2024time,
  title={Time-mmd: Multi-domain multimodal dataset for time series analysis},
  author={Liu, Haoxin and Xu, Shangqing and Zhao, Zhiyuan and Kong, Lingkai and Prabhakar Kamarthi, Harshavardhan and Sasanur, Aditya and Sharma, Megha and Cui, Jiaming and Wen, Qingsong and Zhang, Chao and others},
  journal={Advances in Neural Information Processing Systems},
  volume={37},
  pages={77888--77933},
  year={2024}
}

@article{xu2024intervention,
  title={Intervention-Aware Forecasting: Breaking Historical Limits from a System Perspective},
  author={Xu, Zhijian and Wang, Hao and Xu, Qiang},
  journal={arXiv preprint arXiv:2405.13522},
  year={2024}
}

@inproceedings{merrill2024language,
  title={Language models still struggle to zero-shot reason about time series},
  author={Merrill, Mike A and Tan, Mingtian and Gupta, Vinayak and Hartvigsen, Thomas and Althoff, Tim},
  booktitle={Findings of the Association for Computational Linguistics: EMNLP 2024},
  pages={3512--3533},
  year={2024}
}

@article{chen2025mtbench,
  title={Mtbench: A multimodal time series benchmark for temporal reasoning and question answering},
  author={Chen, Jialin and Feng, Aosong and Zhao, Ziyu and Garza, Juan and Nurbek, Gaukhar and Qin, Cheng and Maatouk, Ali and Tassiulas, Leandros and Gao, Yifeng and Ying, Rex},
  journal={arXiv preprint arXiv:2503.16858},
  year={2025}
}

@inproceedings{wang2025chattime,
  title={Chattime: A unified multimodal time series foundation model bridging numerical and textual data},
  author={Wang, Chengsen and Qi, Qi and Wang, Jingyu and Sun, Haifeng and Zhuang, Zirui and Wu, Jinming and Zhang, Lei and Liao, Jianxin},
  booktitle={Proceedings of the AAAI Conference on Artificial Intelligence},
  volume={39},
  number={12},
  pages={12694--12702},
  year={2025}
}

@article{hurst2024gpt,
  title={Gpt-4o system card},
  author={Hurst, Aaron and Lerer, Adam and Goucher, Adam P and Perelman, Adam and Ramesh, Aditya and Clark, Aidan and Ostrow, AJ and Welihinda, Akila and Hayes, Alan and Radford, Alec and others},
  journal={arXiv preprint arXiv:2410.21276},
  year={2024}
}

@misc{gemini25flash,
  title        = {Gemini 2.5 Flash},
  author       = {{Google DeepMind}},
  year         = {2024},
  howpublished = {\url{https://ai.google.dev/gemini-api/docs/models/gemini}},
  note         = {Accessed 2025}
}

@misc{claude37sonnet,
  title        = {Claude 3.7 Sonnet Model Card},
  author       = {{Anthropic}},
  year         = {2024},
  howpublished = {\url{https://www.anthropic.com/claude}},
  note         = {Accessed 2025}
}

@article{liu2024deepseek,
  title={Deepseek-v3 technical report},
  author={Liu, Aixin and Feng, Bei and Xue, Bing and Wang, Bingxuan and Wu, Bochao and Lu, Chengda and Zhao, Chenggang and Deng, Chengqi and Zhang, Chenyu and Ruan, Chong and others},
  journal={arXiv preprint arXiv:2412.19437},
  year={2024}
}

@article{bai2023qwen,
  title={Qwen technical report},
  author={Bai, Jinze and Bai, Shuai and Chu, Yunfei and Cui, Zeyu and Dang, Kai and Deng, Xiaodong and Fan, Yang and Ge, Wenbin and Han, Yu and Huang, Fei and others},
  journal={arXiv preprint arXiv:2309.16609},
  year={2023}
}

@article{feng2025telecomts,
  title={TelecomTS: A Multi-Modal Observability Dataset for Time Series and Language Analysis},
  author={Feng, Austin and Varvarigos, Andreas and Panitsas, Ioannis and Fernandez, Daniela and Wei, Jinbiao and Guo, Yuwei and Chen, Jialin and Maatouk, Ali and Tassiulas, Leandros and Ying, Rex},
  journal={arXiv preprint arXiv:2510.06063},
  year={2025}
}

@article{lubba2019catch22,
  title={catch22: CAnonical Time-series CHaracteristics: Selected through highly comparative time-series analysis},
  author={Lubba, Carl H and Sethi, Sarab S and Knaute, Philip and Schultz, Simon R and Fulcher, Ben D and Jones, Nick S},
  journal={Data mining and knowledge discovery},
  volume={33},
  number={6},
  pages={1821--1852},
  year={2019},
  publisher={Springer}
}

@article{garza2025timecopilot,
  title={TimeCopilot},
  author={Garza, Azul and Rosillo, Ren{\'e}e},
  journal={arXiv preprint arXiv:2509.00616},
  year={2025}
}

@article{liu2025ts,
  title={TS-Agent: A Time Series Reasoning Agent with Iterative Statistical Insight Gathering},
  author={Liu, Penghang and Fons, Elizabeth and Vyetrenko, Svitlana and Borrajo, Daniel and Potluru, Vamsi and Veloso, Manuela},
  journal={arXiv preprint arXiv:2510.07432},
  year={2025}
}

@inproceedings{lee2025timecap,
  title={Timecap: Learning to contextualize, augment, and predict time series events with large language model agents},
  author={Lee, Geon and Yu, Wenchao and Shin, Kijung and Cheng, Wei and Chen, Haifeng},
  booktitle={Proceedings of the AAAI Conference on Artificial Intelligence},
  volume={39},
  number={17},
  pages={18082--18090},
  year={2025}
}

@article{wu2025out,
  title={Out-of-Distribution Generalization in Time Series: A Survey},
  author={Wu, Xin and Teng, Fei and Li, Xingwang and Zhang, Ji and Li, Tianrui and Duan, Qiang},
  journal={arXiv preprint arXiv:2503.13868},
  year={2025}
}

@article{menchetti2021estimating,
  title={Estimating the causal effect of an intervention in a time series setting: the C-ARIMA approach},
  author={Menchetti, Fiammetta and Cipollini, Fabrizio and Mealli, Fabrizia},
  journal={arXiv preprint arXiv:2103.06740},
  year={2021}
}

@article{shchur2025fev,
  title={fev-bench: A realistic benchmark for time series forecasting},
  author={Shchur, Oleksandr and Ansari, Abdul Fatir and Turkmen, Caner and Stella, Lorenzo and Erickson, Nick and Guerron, Pablo and Bohlke-Schneider, Michael and Wang, Yuyang},
  journal={arXiv preprint arXiv:2509.26468},
  year={2025}
}

@article{chang2025time,
  title={Time-IMM: A Dataset and Benchmark for Irregular Multimodal Multivariate Time Series},
  author={Chang, Ching and Hwang, Jeehyun and Shi, Yidan and Wang, Haixin and Peng, Wen-Chih and Chen, Tien-Fu and Wang, Wei},
  journal={arXiv preprint arXiv:2506.10412},
  year={2025}
}


\newpage
\appendix
\section{More details}


\paragraph{Task Counts per Tier.}
We report the task counts for each tier (T1--T4) across all datasets. For \textbf{MIMIC}, 47 samples are constructed per tier, yielding 188/141/239/141 multiple-choice questions for T1--T4 respectively; T2 and T4 additionally include 47 forecasting samples covering 282 target series. For \textbf{PSML}, each tier contains 50 samples, with 200/150/250/150 multiple-choice questions for T1--T4 and 50 forecasting samples in T2 and T4. For \textbf{Causal Chambers}, each tier also includes 50 samples, with 150/150/250/150 multiple-choice questions for T1--T4 and 50 forecasting samples in T2 and T4. For \textbf{FreshRetailNet}, 44 samples are used per tier, producing 176/132/176/132 multiple-choice questions for T1--T4, with 44 forecasting samples in T2 and T4.

\paragraph{Input Length Analysis.}Table~\ref{tab:token_length_by_dataset} further reports the average input token length of tasks T1--T4 across datasets, computed using the \texttt{gpt-4o} tokenizer, illustrating the substantial contextual and temporal complexity of the benchmark.

\begin{table}[htbp]
\centering
\footnotesize
\setlength{\tabcolsep}{6pt}
\renewcommand{\arraystretch}{1.15}
\begin{tabular}{lcccc}
\toprule
\textbf{Dataset} & \textbf{T1} & \textbf{T2} & \textbf{T3} & \textbf{T4} \\
\midrule
FreshRetailNet   & 23{,}756 & 57{,}895 & 23{,}968 & 58{,}162 \\
PSML             & 4{,}507  & 78{,}397 & 12{,}374 & 74{,}980 \\
MIMIC            & 629      & 6{,}557  & 2{,}078  & 4{,}439  \\
Causal Chambers  & 6{,}963  & 56{,}359 & 26{,}565 & 52{,}828 \\
\bottomrule
\end{tabular}
\caption{Average input token length (computed using the \texttt{gpt-4o} tokenizer) for tasks T1--T4 across different datasets.}
\label{tab:token_length_by_dataset}
\vspace{-10pt}
\end{table}
\paragraph{Conceptual Organization.}
Table~\ref{tab:task_families} summarizes the conceptual organization of T1–T4.
The four task families form a $2\times2$ decomposition that separates historical understanding from future prediction, and isolates how the presence of context and events fundamentally changes the nature of temporal reasoning.

Each task family isolates a different aspect of temporal intelligence by selectively controlling the availability of future targets and contextual information.
Importantly, performance on later task families does not subsume earlier ones: success in prediction does not imply understanding, and access to context does not guarantee effective reasoning.

\begin{table*}[htbp]
\centering
\caption{Conceptual decomposition of temporal competencies in TemporalBench.}
\label{tab:task_families}
\begin{tabular}{lcc}
\toprule
 & \textbf{No Context / Events} & \textbf{With Context / Events} \\
\midrule
\textbf{Historical Focus} &
\textbf{T1: Structural Understanding} &
\textbf{T3: Contextual Temporal Reasoning} \\
 &
Interpret intrinsic temporal properties &
Ground temporal patterns in domain semantics \\
 &
(trend, volatility, seasonality, anomalies) &
and reason about conditions and comparisons \\
\midrule
\textbf{Future Focus} &
\textbf{T2: Pure Temporal Prediction} &
\textbf{T4: Event-Conditioned Prediction} \\
 &
Extrapolate future behavior from &
Reason about how future outcomes change \\
 &
historical temporal signals alone &
under specified upcoming events \\
\bottomrule
\end{tabular}
\end{table*}

\section{Related Work}
\subsection{Time-Series Forecasting Benchmarks}

A number of large-scale benchmarks evaluate forecasting and time-series models across diverse domains, primarily through numerical prediction or analysis objectives.
TFB~\cite{qiu2024tfb} provides a unified framework for fair comparison of statistical, machine learning, and deep learning forecasters.
GIFT-Eval~\cite{aksu2024gifteval} focuses on standardized evaluation protocols for time-series foundation models under pretraining and finetuning splits, and Chronos~\cite{ansari2024chronos} evaluates forecasting models on a large collection of real-world time series datasets, including both in-domain and zero-shot benchmark settings.
Beyond forecasting, several benchmarks target specific time-series analysis capabilities, such as TSI-Bench~\cite{du2024tsi}, TSB-AD~\cite{liu2024elephant}, and Time-MMD~\cite{liu2024time}, which assess performance on analysis- or distribution-oriented tasks over numerical sequences.
More recent efforts begin to incorporate contextual cues into time-series settings, including CiK~\cite{williams2024context} and TGTSF~\cite{xu2024intervention}, but they still frame evaluation largely as single-task numerical analysis without explicit reasoning-oriented objectives.

Meanwhile, a few benchmarks explore the intersection of language models and numerical time-series by combining question answering with time-series inputs, such as LLM TS Struggle~\cite{merrill2024language}, MTBench~\cite{chen2025mtbench}, and ChatTime~\cite{wang2025chattime}.
While these benchmarks move beyond pure forecasting, they typically do not explicitly incorporate external contextual information or provide a unified multi-task taxonomy to disentangle complementary aspects of temporal competence.
Overall, existing time-series benchmarks are largely optimized for numerical performance, motivating our effort to move beyond prediction error toward understanding- and context-aware temporal reasoning.

\subsection{Temporal Reasoning Benchmarks for LLMs}

A separate line of work evaluates temporal reasoning in large language models using purely textual problems, focusing on symbolic, event-based, and commonsense temporal understanding.
Benchmarks such as TimeBench~\cite{chu2024timebench} and other temporal reasoning evaluations probe abilities such as ordering events, reasoning about duration, and inferring causal-temporal relations from text.
These datasets highlight important deficiencies of LLMs in temporal reasoning, but they omit real numerical time-series signals and do not capture the numerical temporal dynamics central to forecasting and many real-world decision-making settings.

Recent multi-modal benchmarks begin to bridge this gap by coupling numerical time-series with language supervision and downstream reasoning tasks.
For example, TelecomTS~\cite{feng2025telecomts} introduces a large-scale observability benchmark derived from real-world telecommunications networks, supporting tasks such as anomaly detection, root-cause analysis, and question answering over time-series with associated textual context.
Such efforts further motivate the need for benchmarks that jointly evaluate numerical time-series understanding and contextual reasoning, which our work addresses.

\subsection{Agentic and LLM-Based Methods for Time-Series}

Recent work has begun to explore agentic or large language model (LLM)–driven approaches for time-series analysis and forecasting.
TimeSeriesScientist~\cite{zhao2025timeseriesscientist} proposes an agent-based framework that decomposes univariate forecasting into stages such as data inspection, model selection, and result interpretation, demonstrating the potential of agentic workflows for automating parts of the forecasting pipeline.
Other studies investigate the capabilities and limitations of directly applying LLMs to time-series tasks, including numerical forecasting, trend interpretation, and zero-shot reasoning over serialized temporal inputs~\cite{jin2024whatcanllm}.

More recent work extends this line of research by introducing agentic architectures that combine LLM reasoning with time-series–specific tools or models.
TimeCAP~\cite{lee2025timecap} leverages LLM agents to contextualize and semantically augment time-series data for downstream prediction,
while TS-Agent~\cite{liu2025ts} emphasizes iterative statistical reasoning through tool invocation and evidence accumulation.
From a systems perspective, TimeCopilot~\cite{garza2025timecopilot} presents an open-source agentic forecasting framework that automates model selection, ensembling, and explanation generation.

In parallel, several general-purpose agent frameworks have been proposed to support structured reasoning and coordination among multiple LLM agents.
MetaGPT~\cite{hong2024metagpt} organizes agents according to software engineering roles (e.g., product manager, architect, developer), enabling role-specialized collaboration through explicit communication protocols.
AgentScope~\cite{gao2024agentscope} provides a flexible platform for constructing, orchestrating, and evaluating multi-agent systems, with explicit support for tool use, memory, and interaction logging.
CAMEL~\cite{li2023camel} emphasizes autonomous role-playing and conversational coordination between agents to elicit complex reasoning behaviors without heavy task-specific supervision.
While these frameworks offer powerful abstractions for agent coordination and reasoning, they are largely task-agnostic and do not provide dedicated evaluation settings tailored to time-series understanding.

\section{More Analysis}

\subsection{Error Analysis}
\label{subsec:error_analysis}
\begin{figure}[htbp]
    \centering
 \includegraphics[width=\columnwidth]{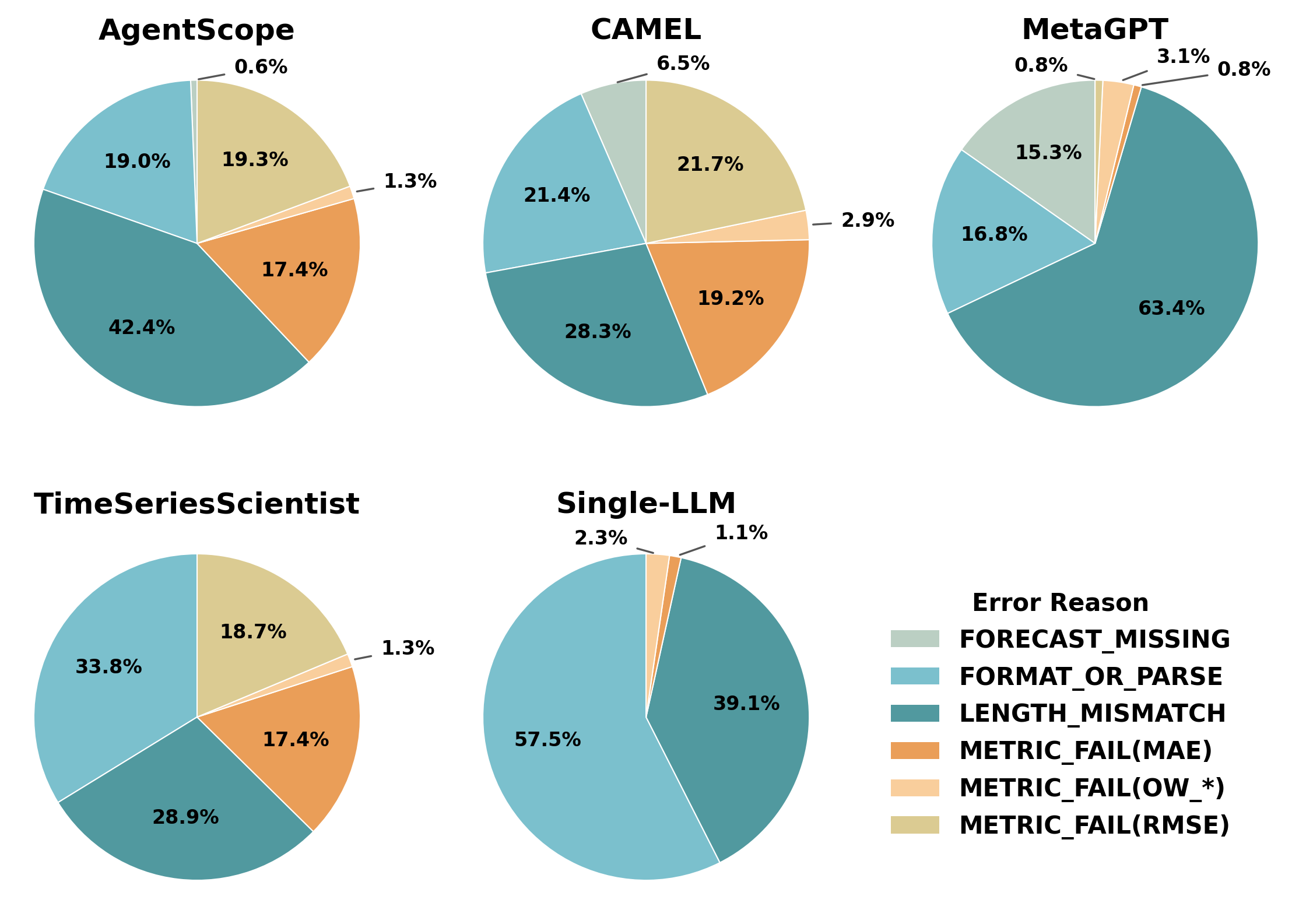}
    \caption{
Distribution of error types across five agents using \texttt{gpt-4o} as the base LLM, aggregated over all datasets and tasks.
Each pie chart corresponds to an agent and shows the proportion of different failure modes.
}

    \label{fig:error_type_distribution}
\end{figure}
Figure~\ref{fig:error_type_distribution} presents an aggregated error analysis grouped by agent, where all systems are instantiated on the same ChatGPT backbone. For clarity, we categorize failures into several broad types: failures to produce valid answers for multiple-choice questions, violations of required output formats, incorrect forecast horizon lengths, and severe numerical abnormalities reflected by extreme error metrics. While these categories correspond to the color-coded groups in the figure, we focus our discussion on their semantic implications rather than their internal labels.

\paragraph{Overall trends}
A dominant observation across all settings is that a large fraction of errors arise from \emph{output control failures} rather than a lack of temporal understanding. In particular, predicting a sequence with an incorrect horizon length is the most frequent failure mode for general-purpose agents. This issue accounts for more than 40\% of AgentScope’s errors, nearly two-thirds of MetaGPT’s errors, and remains substantial even for CAMEL. The single-LLM baseline exhibits a similar pattern, with length violations constituting a large portion of its failures. These results suggest that maintaining strict control over forecast length is a fundamental challenge when LLMs are used to generate structured temporal outputs, regardless of whether an agent framework is employed.

Beyond horizon control, numerical instability emerges as another prominent source of error. For several agents, a considerable number of outputs satisfy basic syntactic constraints yet lead to extremely large MAE or RMSE values. Such failures indicate that models may generate sequences that are formally valid but numerically implausible, for example due to scale drift, unbounded growth, or abrupt regime changes not supported by the input history.

\paragraph{Comparison across agent types}
Despite sharing the same base model, different agent designs exhibit markedly different error profiles. General-purpose agents tend to struggle most with low-level execution constraints. MetaGPT, for example, is overwhelmingly dominated by horizon-length errors, indicating brittle control over structured generation in multi-step workflows. AgentScope and CAMEL show more balanced distributions, but still suffer heavily from a combination of length violations and numerically unstable forecasts, suggesting that orchestration and tool usage alone do not resolve these issues.

The time-series–specific agent (TimeSeriesScientist) displays a qualitatively different pattern. Errors related to selecting invalid options in multiple-choice tasks are largely mitigated, indicating better alignment with task semantics. However, this advantage comes with a shift toward formatting violations and numerical anomalies. In other words, while domain-specific reasoning helps avoid some semantic errors, it does not guarantee adherence to strict output schemas nor robustness in numerical generation.

The single-LLM baseline presents the most extreme form of output unreliability. A majority of its failures stem from producing outputs that either do not conform to the required format or do not match the expected horizon length. Metric-related failures are comparatively rare, largely because many responses fail before reaching numerical evaluation. This highlights the role of agent scaffolding in filtering out the most basic execution errors, even if deeper issues remain.

\paragraph{Implications for future agent design}
Taken together, these results suggest that the primary bottlenecks for agentic time-series modeling lie not in temporal reasoning itself, but in enforcing execution discipline. Robust agents must treat output constraints as first-class citizens. This includes explicit mechanisms to guarantee horizon length, enforce output schemas, and validate numerical plausibility before accepting a forecast. Without such safeguards, even strong language models and sophisticated agent workflows remain vulnerable to systematic, repeatable failure modes.

Overall, the error analysis reveals that many failures are structural rather than incidental. Addressing these issues requires rethinking agent design around controllability and validation, rather than solely improving temporal representations or reasoning strategies.

\begin{insightbox}
\textbf{Insight.}
A substantial portion of errors across T2 and T4 stem from execution and controllability failures rather than incorrect temporal reasoning. This suggests that improving agentic time-series systems requires treating output constraints and validation as first-class design objectives alongside reasoning capabilities.
\end{insightbox}

\subsection{Length Sensitivity Study}
\label{subsec:length_sensitivity}

\begin{figure}[htbp]
    \centering
 \includegraphics[width=\columnwidth]{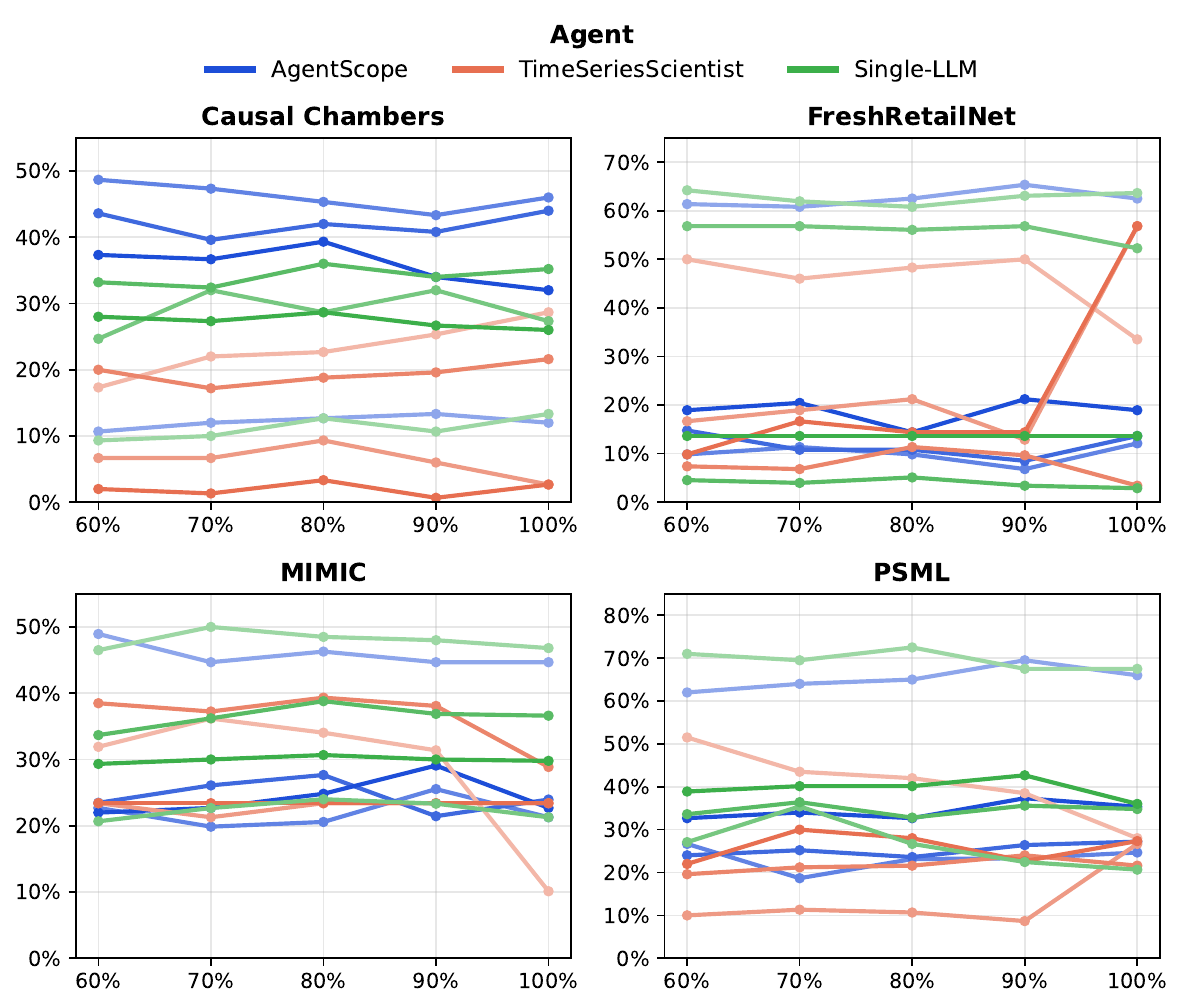}
\caption{
Effect of input time-series length on agent performance across four datasets.
Each subplot corresponds to a dataset, and multi-choice accuracy is reported for different agents under varying input lengths.
Lines with the same color denote the same agent, while darker to lighter shades represent T1--T4 tasks, respectively.
}

    \label{fig:length_sensitivity_by_dataset}
\end{figure}

Figure~\ref{fig:length_sensitivity_by_dataset} studies how truncating the historical input affects performance across datasets and task tiers (T1--T4) for three representative settings: a general-purpose agent (AgentScope), a time-series agent (TimeSeriesScientist), and a single-LLM baseline.
Overall, history length is not a universally monotonic driver of performance.
Its effect varies by domain and tier, and longer inputs can introduce instability rather than steady gains.
This length sensitivity matters in practice because real agent pipelines often face prompt budget limits and must decide whether to include more raw observations, to summarize them, or to retrieve only salient segments.
The results here therefore highlight a trade-off: longer histories may provide more evidence, but they can also dilute key signals and make structured generation harder to control.
In several settings we observe “sweet spots,” where intermediate context performs as well as (or better than) the longest input, suggesting that effective utilization of additional tokens is itself a non-trivial capability.

\paragraph{Dataset-dependent sensitivity is substantial.}
The four datasets respond differently as history increases.
On PSML, AgentScope and the single-LLM baseline are relatively stable and often improve slightly with more context, especially on simpler tasks such as T1, suggesting recurring patterns that remain exploitable under straightforward truncation.
In contrast, FreshRetailNet exhibits weaker and more irregular sensitivity: most tiers fluctuate within a narrow band as the window grows, indicating limited marginal evidence from additional retail history.
MIMIC shows the clearest length-induced instability---moderate extensions can help, but very long histories sometimes degrade sharply, most noticeably for the hardest tier---consistent with long clinical histories introducing noise that is difficult for LLM-driven pipelines to compress and use.
Causal Chambers is more consistent than retail and MIMIC, though performance still varies across tiers, indicating that domain structure, not raw length, primarily governs difficulty.

\paragraph{Task-tier effects reveal diminishing returns and occasional reversals.}
Across datasets, T1 benefits most consistently from longer history, but improvements often saturate once a substantial portion of the series is included.
For AgentScope, the best T1 performance frequently appears at intermediate-to-long histories rather than at the maximum length.
For prediction- and decision-oriented tasks (T2, T4), trends are less stable: moderate gains can appear, yet they rarely persist, and the longest inputs can even reverse performance.
This aligns with the earlier observation that these tiers are limited not only by temporal evidence, but by the agent’s ability to extract relevant signals and maintain controllable, structured outputs as prompts grow.

\paragraph{Differences across agent types highlight distinct failure modes.}
AgentScope is generally robust but relatively insensitive to history length, especially on FreshRetailNet, suggesting that longer raw histories are not automatically converted into stronger temporal reasoning.
The single-LLM baseline shows a clearer sweet spot: moderate lengths often match or outperform both shorter and longer settings, while the longest prompts can degrade performance, reflecting a trade-off between contextual richness and controllability.
TimeSeriesScientist is more bimodal: in some domain--tier combinations it exploits longer histories effectively (e.g., a sharp gain at the longest input on FreshRetailNet for one tier), but it can also drop abruptly elsewhere (e.g., severe degradation at the longest history on MIMIC for a difficult tier).
This suggests that time-series--specific tooling can help when extra history contains relevant structure, but becomes brittle when long contexts include heterogeneous regimes or are not compressed into stable intermediate representations.

\paragraph{Implications for agent design and benchmark use.}
Two implications follow.
First, longer history should not be equated with better context: beyond a moderate length, additional tokens can yield diminishing returns or reversals, so evaluations should not assume that more context is inherently beneficial.
Second, the results motivate length-aware agents.
Instead of truncation alone, systems should add history selection and summarization mechanisms (e.g., retrieving salient subsequences, compressing regimes, extracting structured evidence) to prevent long-input dilution and instability.
From a benchmarking perspective, reporting performance at multiple history lengths remains essential: it reveals whether methods truly leverage extended evidence or rely on short-horizon heuristics, and avoids conclusions driven by a single fixed length.

Overall, the length study suggests that temporal benchmarks should evaluate not only forecasting and reasoning accuracy, but also an agent’s ability to robustly use additional historical evidence under realistic prompt-length and control constraints.
This capability remains only partially and inconsistently developed across current general agents, time-series agents, and single-LLM baselines.

\begin{insightbox}
\textbf{Insight.}
The ability to leverage longer historical context is fragile and task-dependent. Performance saturation and degradation at extended input lengths indicate that temporal reasoning is constrained not only by access to evidence, but by an agent’s ability to select, compress, and control relevant temporal information.
\end{insightbox}

\section{ Pseudocode for Event Injection / Detection and Ground-Truth Labeling}
\label{app:pseudocode}

The section~\ref{sec:benchmark} provides algorithmic descriptions of the dataset transformation pipeline introduced in Section~3, including event injection or detection and rule-based ground-truth generation.
The pseudocode is intended to make the benchmark construction process fully explicit and reproducible, translating the methodological principles discussed in the main text into concrete procedures.
In particular, these algorithms clarify how event boundaries are defined independently of label computation, how robust statistical criteria are used to derive ground truth across task families, and how uncertainty is handled when temporal signals are weak or ambiguous.
Together, they serve as a reference implementation of the benchmark design rather than a simulation of real-world causal dynamics.

\begin{algorithm}[H]
\caption{Event Generation: Injection \textbf{or} Detection}
\label{alg:event_gen}
\begin{algorithmic}[1]
\Require Time series dataset $\mathcal{D} = \{(x^{(i)}_{1:T_i}, meta^{(i)})\}_{i=1}^N$, randomness seed $s$, injection probability $p_{inj}$, injection magnitudes / patterns set $\mathcal{P}$, change-point detector $CP(\cdot)$
\Ensure For each series index $i$, event time $t^{(i)}_{evt}$ and history/future split $(H^{(i)}, F^{(i)})$
\State set random seed $s$
\For{each time series $i=1\ldots N$}
    \State $x \leftarrow x^{(i)}_{1:T_i}$
    \If{$meta^{(i)}$ contains reliable event annotations}
        \State $t_{evt} \leftarrow$ recorded event time (choose one if multiple)
        \State mark $mode \leftarrow$ \texttt{DETECTED}
    \Else
        \State $cp\_candidates \leftarrow CP(x)$ \Comment{e.g., ruptures, PELT, or domain-specific detector}
        \If{$cp\_candidates \neq \varnothing \land random() < p_{det}$}
            \State $t_{evt} \leftarrow$ select candidate (e.g., largest magnitude)
            \State mark $mode \leftarrow$ \texttt{DETECTED}
        \Else
            \If{random() $<$ $p_{inj}$}
                \State $t_{evt} \leftarrow$ select injection time (e.g., uniform over valid window)
                \State $pattern \leftarrow$ sample($\mathcal{P}$)
                \State $x \leftarrow$ InjectEffect$(x, t_{evt}, pattern)$
                \State mark $mode \leftarrow$ \texttt{INJECTED}
            \Else
                \State \textbf{skip} this series (no event defined)
            \EndIf
        \EndIf
    \EndIf
    \State Define history window $H^{(i)} = x_{1:t_{evt}}$ and future window $F^{(i)} = x_{t_{evt}+1 : t_{evt}+h}$, where $h$ is forecast horizon
    \State Save $(H^{(i)}, F^{(i)}, t_{evt}, mode)$
\EndFor
\State \Return $\{(H^{(i)}, F^{(i)}, t_{evt}, mode)\}_{i}$
\end{algorithmic}
\end{algorithm}

\vspace{6pt}
\noindent\textbf{Notes / Recommended defaults:}
\begin{itemize}
  \item $p_{inj}$: injection probability for datasets lacking events, e.g. 0.3--0.5. Use consistent value across experiments.
  \item $\mathcal{P}$ (injection patterns): coarse patterns only (shift in level, scale change, temporary spike, seasonality phase shift). Avoid parameterizing patterns to produce deterministic labels.
  \item $CP(\cdot)$: preferred detectors include PELT or Bayesian change-point detectors; tune sensitivity but report chosen hyperparameters.
  \item Choose injection times away from boundaries (e.g., avoid first/last 10\% of series) to ensure adequate history and future support.
  \item Log and publish the RNG seed(s) and injection choices to ensure reproducibility.
\end{itemize}

\bigskip

\begin{algorithm}[H]
\caption{Rule-based Ground-Truth Labeling}
\label{alg:label_gen}
\begin{algorithmic}[1]
\Require Partitioned series $(H, F)$, minimum sample support $n_{min}$, effect-size thresholds $\tau_{level}, \tau_{vol}$, estimators: median, MAD, IQR, Theil--Sen
\Ensure Labels for T1/T2/T3/T4 tasks; possibly \texttt{UNCERTAIN} or \texttt{INCONCLUSIVE}
\State Compute robust estimates on $H$: $\tilde{\mu}_H \leftarrow median(H)$; $\tilde{\sigma}_H \leftarrow MAD(H)$; seasonal features if applicable
\State Compute robust estimates on $F$: $\tilde{\mu}_F \leftarrow median(F)$; $\tilde{\sigma}_F \leftarrow MAD(F)$
\State $n_H \leftarrow$ length$(H)$; $n_F \leftarrow$ length$(F)$
\If{$n_H < n_{min} \lor n_F < n_{min}$}
    \State Mark labels as \texttt{INCONCLUSIVE}
    \State \Return
\EndIf

\State \textbf{Level change:} $d_{level} \leftarrow (\tilde{\mu}_F - \tilde{\mu}_H)/\max(|\tilde{\mu}_H|, \epsilon)$
\State \textbf{Volatility change:} $d_{vol} \leftarrow (\tilde{\sigma}_F - \tilde{\sigma}_H)/\max(\tilde{\sigma}_H, \epsilon)$
\State \textbf{Effect sizes:} compute effect size (e.g., Cliff's delta or rank-biserial) for distributions $H$ vs $F$
\State \textbf{Significance criterion:} $support = (|d_{level}| > \tau_{level}) \lor (|d_{vol}| > \tau_{vol}) \lor (\text{effect size} > \tau_{es})$

\If{support is False}
    \State Assign qualitative label = \texttt{UNCERTAIN}
\Else
    \State Assign qualitative labels:
    \If{$d_{level} > \tau_{level}$}
        \State label level = \texttt{INCREASE}
    \ElsIf{$d_{level} < -\tau_{level}$}
        \State label level = \texttt{DECREASE}
    \Else
        \State label level = \texttt{NO\_CHANGE}
    \EndIf
    \If{$d_{vol} > \tau_{vol}$}
        \State label volatility = \texttt{INCREASE}
    \ElsIf{$d_{vol} < -\tau_{vol}$}
        \State label volatility = \texttt{DECREASE}
    \Else
        \State label volatility = \texttt{NO\_CHANGE}
    \EndIf
\EndIf

\State \textbf{T1 labels (history-only):}
\State compute windowed comparisons on $H$ (early vs late) using relative median and dispersion ratios $\to$ trend / seasonality / anomaly labels (use Theil--Sen for slopes)
\State apply robust z-scores for anomaly detection: $z_t = (x_t - median(H))/MAD(H)$; mark extreme outliers if $|z_t| > z_{thresh}$

\State \textbf{T2 / T4 numeric ground truth:} store observed $F$ values as numeric forecast targets

\State \textbf{T3 labels (contextual reasoning):}
\State for candidate contextual queries (e.g., difference between subgroups), compute effect size and ensure $n_{support} \ge n_{min}$; only retain queries with effect size $> \tau_{es}$

\State \textbf{Finalize:} attach metadata: $(labels, support\_metrics, mode)$ where mode indicates \texttt{DETECTED} or \texttt{INJECTED}

\State \Return labels and metadata
\end{algorithmic}
\end{algorithm}

\vspace{6pt}
\noindent\textbf{Labeling constants and recommended defaults:}
\begin{itemize}
  \item $n_{min}$ (minimum samples): 10--30 depending on frequency (daily vs hourly)
  \item $\tau_{level}$ (relative median change): 0.10 (10\% relative change) as baseline; report sensitivity analyses
  \item $\tau_{vol}$ (relative MAD change): 0.10
  \item $\tau_{es}$ (effect size threshold): 0.2 (small-to-medium)
  \item $z_{thresh}$ (robust z-score for anomalies): 3.5
  \item $\epsilon$: small constant to avoid divide-by-zero (e.g., $10^{-6}$)
\end{itemize}

\vspace{6pt}
\noindent\textbf{Reproducibility and anti-shortcut measures:}
\begin{itemize}
  \item Random seeds for injection/detection sampling are fixed and recorded. Publish seeds and the exact sampled injection patterns in the artifact.
  \item Use multiple injection patterns and random magnitudes within a coarse range to prevent models from learning deterministic templates.
  \item Ensure all label decisions are computed only from $H$ and $F$ arrays: explicitly do not use the textual context or injected pattern metadata in any labeling rule.
  \item Record support metrics (effect sizes, sample counts) with each label so that downstream analyses can filter or weight by label confidence.
\end{itemize}

\vspace{6pt}
\noindent\textbf{Design rationale (concise):}
\begin{itemize}
  \item The pipeline favors \emph{controlled, interpretable evaluation signals} over exact simulation fidelity: event injection is intended as a way to create regime splits for probing event-aware reasoning, not as a synthetic causal generator.
  \item Robust estimators (median/MAD/IQR/Theil--Sen) are used to make labels stable across domains and data scales.
  \item Uncertain / Inconclusive labels avoid forcing spurious ground truth when data support is weak.
\end{itemize}

\section{Illustrative Task Examples Across Domains}
\begin{figure}[htbp]
\centering
\begin{tcolorbox}[
  colback=gray!6,
  colframe=black!20,
  boxrule=0.5pt,
  arc=3pt,
  left=6pt,right=6pt,top=6pt,bottom=6pt
]
\textbf{Dataset:} PSML \\

\emph{Shared time-series context:}
\begin{itemize}[leftmargin=1.1em,itemsep=0pt,topsep=0pt,parsep=0pt,partopsep=0pt]
  \item Target variable: \texttt{load\_power}
  \item History length: 336 \quad | \quad Future length: 168
  \item History: 0.805, 0.779, 0.770, 0.792, \ldots, 0.975, 0.893, 0.836
  \item Future: 0.720, 0.689, 0.703, 0.709, \ldots, 0.431, 0.509, 0.427
  \item Covariates:
  \texttt{avg\_temperature}, \texttt{time\_position\_in\_day},
  Temperature, Wind Speed, GHI, DHI, DNI,
  Relative Humidity, Solar Zenith Angle
\end{itemize}

\hrule\vspace{4pt}

\textbf{T1}
\begin{itemize}[leftmargin=1.1em,itemsep=0pt,topsep=0pt,parsep=0pt,partopsep=0pt]
  \item Trend (\textit{upward / downward / constant}): \textbf{constant}
  \item Volatility (\textit{increased / decreased / constant}): \textbf{constant}
  \item Seasonality (\textit{fixed / shifting / none}): \textbf{fixed}
  \item Outliers (\textit{sudden\_spike / level\_shift / stable}): \textbf{sudden\_spike}
\end{itemize}

\vspace{3pt}\hrule\vspace{4pt}

\textbf{T2}

\emph{Forecasting subtask:}
Predict future values of the target variable given the shared historical context.

\emph{MCQ subtask:}
\begin{itemize}[leftmargin=1.1em,itemsep=0pt,topsep=0pt,parsep=0pt,partopsep=0pt]
  \item Future vs. History
  (\textit{Higher / Lower / Similar / Uncertain}): \textbf{Lower}
  \item Volatility Change
  (\textit{increased / decreased / constant / Uncertain}): \textbf{increased}
  \item Seasonality Shift
  (\textit{fixed / shifting / no / Uncertain}): \textbf{fixed}
\end{itemize}

\vspace{3pt}\hrule\vspace{4pt}

\textbf{T3} \\
\emph{Question:} A sharp nighttime load rise is observed. Does it coincide with a temperature drop? \\
\emph{Options:} \textit{Yes / No / Uncertain} \\
\textbf{Answer:} \textbf{Yes}

\vspace{3pt}\hrule\vspace{4pt}

\textbf{T4}

\emph{Event description:}
A sharp nighttime load rise accompanied by a temperature drop is observed in the historical window.

\emph{Forecasting subtask:}
Predict future values conditioned on the shared historical context and the above event.

\emph{MCQ subtask:}
\begin{itemize}[leftmargin=1.1em,itemsep=0pt,topsep=0pt,parsep=0pt,partopsep=0pt]
  \item Future vs. History
  (\textit{Higher / Lower / Similar / Uncertain}): \textbf{Lower}
  \item Volatility Change
  (\textit{increased / decreased / constant / Uncertain}): \textbf{increased}
  \item Seasonality Shift
  (\textit{fixed / shifting / no / Uncertain}): \textbf{fixed}
\end{itemize}

\end{tcolorbox}
\caption{A full example instance (PSML) showing tasks T1--T4 in TemporalBench. All tiers share the same underlying time-series context, while T4 is additionally conditioned on an explicit event description.}
\label{fig:appendix_psml_one_example}
\end{figure}

\begin{figure}[htbp]
\centering
\begin{tcolorbox}[
  colback=gray!6,
  colframe=black!20,
  boxrule=0.5pt,
  arc=3pt,
  left=6pt,right=6pt,top=6pt,bottom=6pt
]
\textbf{Dataset:} MIMIC \\

\emph{Shared time-series context:}
\begin{itemize}[leftmargin=1.1em,itemsep=0pt,topsep=0pt,parsep=0pt,partopsep=0pt]
  \item Reference variable: \texttt{heart\_rate}
  \item History length: 50 \quad | \quad Future length: 29
  \item History: 80.0, 82.0, 82.0, 82.0, \ldots, 92.0, 92.0, 91.0
  \item Future: 91.0, 91.0, 80.0, 85.0, \ldots, 84.0, 84.0, 86.33
  \item Covariates (history):
  \texttt{temperature\_c}, \texttt{resp\_rate}, \texttt{spo2},
  \texttt{sbp}, \texttt{dbp}, \texttt{time\_position\_in\_day}
  \item Covariates (future):
  \texttt{time\_position\_in\_day}
\end{itemize}

\hrule\vspace{4pt}

\textbf{T1}
\begin{itemize}[leftmargin=1.1em,itemsep=0pt,topsep=0pt,parsep=0pt,partopsep=0pt]
  \item Trend (\textit{upward / downward / constant}): \textbf{constant}
  \item Volatility (\textit{increased / decreased / constant}): \textbf{increased}
  \item Seasonality (\textit{fixed / shifting / none}): \textbf{shifting}
  \item Outliers (\textit{sudden\_spike / level\_shift / stable}): \textbf{sudden\_spike}
\end{itemize}

\vspace{3pt}\hrule\vspace{4pt}

\textbf{T2}

\emph{Forecasting subtask:}
Predict future \emph{multivariate physiological time-series} trajectories given the shared historical context, with heart rate serving as the primary reference variable.

\emph{MCQ subtask:}
\begin{itemize}[leftmargin=1.1em,itemsep=0pt,topsep=0pt,parsep=0pt,partopsep=0pt]
  \item Future vs. History
  (\textit{Higher / Lower / Similar / Uncertain}): \textbf{Uncertain}
  \item Volatility Change
  (\textit{increased / decreased / constant / Uncertain}): \textbf{decreased}
  \item Seasonality Shift
  (\textit{fixed / shifting / no / Uncertain}): \textbf{shifting}
\end{itemize}

\vspace{3pt}\hrule\vspace{4pt}

\textbf{T3} \\
\emph{Question:} During fever episodes (Temp $\ge$ 37.2$^\circ$C), is heart rate higher than normal? \\
\emph{Options:} \textit{Higher / Lower / Similar / Uncertain} \\
\textbf{Answer:} \textbf{Higher}

\vspace{3pt}\hrule\vspace{4pt}

\textbf{T4}

\emph{Event description:}
Fever episodes are observed when body temperature reaches or exceeds 37.2$^\circ$C.

\emph{Forecasting subtask:}
Predict future \emph{multivariate physiological time-series} trajectories conditioned on the shared historical context and the fever event, with heart rate serving as the primary reference variable.

\emph{MCQ subtask:}
\begin{itemize}[leftmargin=1.1em,itemsep=0pt,topsep=0pt,parsep=0pt,partopsep=0pt]
  \item Future vs. History
  (\textit{Higher / Lower / Similar / Uncertain}): \textbf{Uncertain}
  \item Volatility Change
  (\textit{increased / decreased / constant / Uncertain}): \textbf{decreased}
  \item Seasonality Shift
  (\textit{fixed / shifting / no / Uncertain}): \textbf{shifting}
\end{itemize}

\end{tcolorbox}
\caption{A full example instance (MIMIC) illustrating tasks T1--T4 in TemporalBench. All tiers share the same underlying multivariate physiological time-series context, while T4 is additionally conditioned on a clinical event (fever).}
\label{fig:appendix_mimic_one_example}
\end{figure}

\begin{figure}[htbp]
\centering
\begin{tcolorbox}[
  colback=gray!6,
  colframe=black!20,
  boxrule=0.5pt,
  arc=3pt,
  left=6pt,right=6pt,top=6pt,bottom=6pt
]
\textbf{Dataset:} FreshRetailNet \\

\emph{Shared time-series context:}
\begin{itemize}[leftmargin=1.1em,itemsep=0pt,topsep=0pt,parsep=0pt,partopsep=0pt]
  \item Target variable: \texttt{sales\_censored}
  \item History length: 480 \quad | \quad Future length: 112
  \item History: 0.0, 0.2, 0.0, 0.3, \ldots, 0.0, 0.0, 0.0
  \item Future: 0.091, 0.0, 0.817, 0.465, \ldots, 0.336, 0.0, 0.100
  \item Covariates:
  \texttt{discount}, \texttt{holiday\_flag}, \texttt{precipitation},
  \texttt{avg\_temperature}, \texttt{time\_position\_in\_day}
\end{itemize}

\hrule\vspace{4pt}

\textbf{T1}
\begin{itemize}[leftmargin=1.1em,itemsep=0pt,topsep=0pt,parsep=0pt,partopsep=0pt]
  \item Trend (\textit{upward / downward / constant}): \textbf{constant}
  \item Volatility (\textit{increased / decreased / constant}): \textbf{constant}
  \item Seasonality (\textit{fixed / shifting / none}): \textbf{fixed}
  \item Outliers (\textit{sudden\_spike / level\_shift / stable}): \textbf{sudden\_spike}
\end{itemize}

\vspace{3pt}\hrule\vspace{4pt}

\textbf{T2}

\emph{Forecasting subtask:}
Predict future retail sales values given the shared historical context.

\emph{MCQ subtask:}
\begin{itemize}[leftmargin=1.1em,itemsep=0pt,topsep=0pt,parsep=0pt,partopsep=0pt]
  \item Future vs. History
  (\textit{Higher / Lower / Similar / Uncertain}): \textbf{Uncertain}
  \item Volatility Change
  (\textit{increased / decreased / constant / Uncertain}): \textbf{Uncertain}
  \item Seasonality Shift
  (\textit{fixed / shifting / no / Uncertain}): \textbf{fixed}
\end{itemize}

\vspace{3pt}\hrule\vspace{4pt}

\textbf{T3} \\
\emph{Question:} Under high-discount periods, is the peak-to-median ratio higher than under low or no discount? \\
\emph{Options:} \textit{Yes / No / Uncertain} \\
\textbf{Answer:} \textbf{No}

\vspace{3pt}\hrule\vspace{4pt}

\textbf{T4}

\emph{Event description:}
High-discount periods are observed in the historical window.

\emph{Forecasting subtask:}
Predict future retail sales values conditioned on the shared historical context and the discount event.

\emph{MCQ subtask:}
\begin{itemize}[leftmargin=1.1em,itemsep=0pt,topsep=0pt,parsep=0pt,partopsep=0pt]
  \item Future vs. History
  (\textit{Higher / Lower / Similar / Uncertain}): \textbf{Uncertain}
  \item Volatility Change
  (\textit{increased / decreased / constant / Uncertain}): \textbf{Uncertain}
  \item Seasonality Shift
  (\textit{fixed / shifting / no / Uncertain}): \textbf{fixed}
\end{itemize}

\end{tcolorbox}
\caption{A full example instance (FreshRetailNet) illustrating tasks T1--T4 in TemporalBench. All tiers share the same underlying retail time-series context, while T4 is additionally conditioned on a promotional event (high discount).}
\label{fig:appendix_freshretailnet_one_example}
\end{figure}

\begin{figure}[htbp]
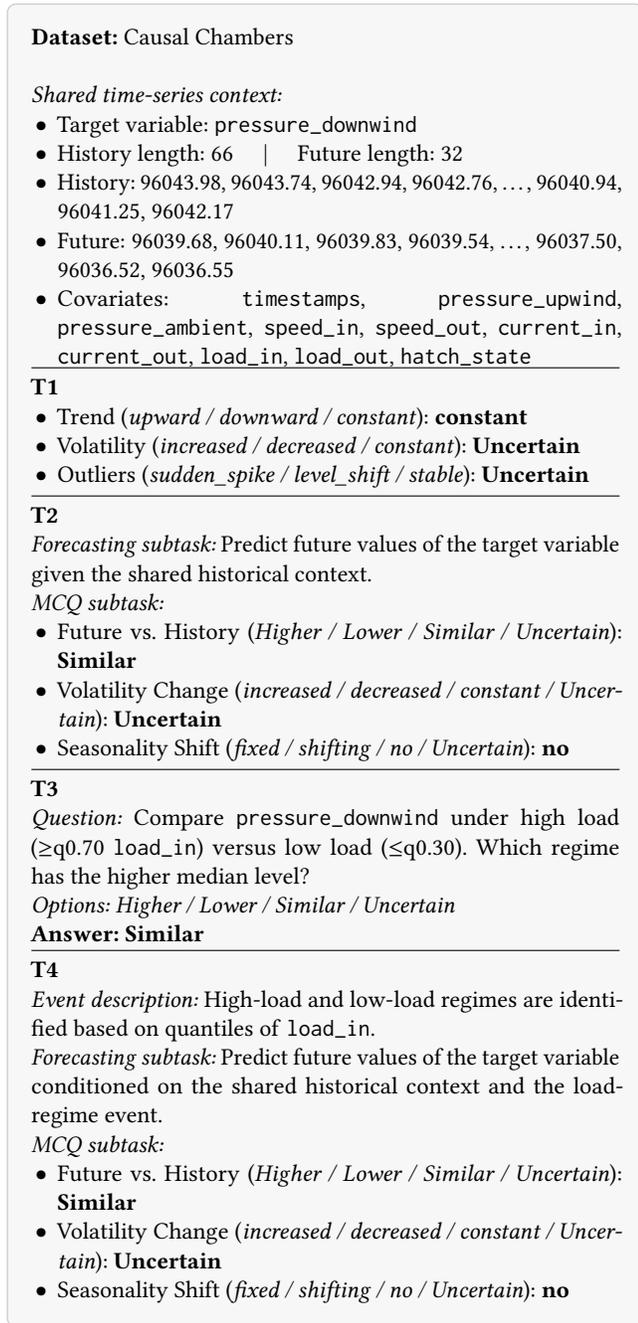

\centering
\begin{tcolorbox}[
  colback=gray!6,
  colframe=black!20,
  boxrule=0.5pt,
  arc=3pt,
  left=6pt,right=6pt,top=6pt,bottom=6pt
]
\textbf{Dataset:} Causal Chambers \\

\emph{Shared time-series context:}
\begin{itemize}[leftmargin=1.1em,itemsep=0pt,topsep=0pt,parsep=0pt,partopsep=0pt]
  \item Target variable: \texttt{pressure\_downwind}
  \item History length: 66 \quad | \quad Future length: 32
  \item History: 96043.98, 96043.74, 96042.94, 96042.76, \ldots, 96040.94, 96041.25, 96042.17
  \item Future: 96039.68, 96040.11, 96039.83, 96039.54, \ldots, 96037.50, 96036.52, 96036.55
  \item Covariates:
  \texttt{timestamps}, \texttt{pressure\_upwind}, \texttt{pressure\_ambient},
  \texttt{speed\_in}, \texttt{speed\_out}, \texttt{current\_in}, \texttt{current\_out},
  \texttt{load\_in}, \texttt{load\_out}, \texttt{hatch\_state}
\end{itemize}

\hrule\vspace{4pt}

\textbf{T1}
\begin{itemize}[leftmargin=1.1em,itemsep=0pt,topsep=0pt,parsep=0pt,partopsep=0pt]
  \item Trend (\textit{upward / downward / constant}): \textbf{constant}
  \item Volatility (\textit{increased / decreased / constant}): \textbf{Uncertain}
  \item Outliers (\textit{sudden\_spike / level\_shift / stable}): \textbf{Uncertain}
\end{itemize}

\vspace{3pt}\hrule\vspace{4pt}

\textbf{T2}

\emph{Forecasting subtask:}
Predict future values of the target variable given the shared historical context.

\emph{MCQ subtask:}
\begin{itemize}[leftmargin=1.1em,itemsep=0pt,topsep=0pt,parsep=0pt,partopsep=0pt]
  \item Future vs. History
  (\textit{Higher / Lower / Similar / Uncertain}): \textbf{Similar}
  \item Volatility Change
  (\textit{increased / decreased / constant / Uncertain}): \textbf{Uncertain}
  \item Seasonality Shift
  (\textit{fixed / shifting / no / Uncertain}): \textbf{no}
\end{itemize}

\vspace{3pt}\hrule\vspace{4pt}

\textbf{T3} \\
\emph{Question:} Compare \texttt{pressure\_downwind} under high load ($\ge$q0.70 \texttt{load\_in}) versus low load ($\le$q0.30). Which regime has the higher median level? \\
\emph{Options:} \textit{Higher / Lower / Similar / Uncertain} \\
\textbf{Answer:} \textbf{Similar}

\vspace{3pt}\hrule\vspace{4pt}

\textbf{T4}

\emph{Event description:}
High-load and low-load regimes are identified based on quantiles of \texttt{load\_in}.

\emph{Forecasting subtask:}
Predict future values of the target variable conditioned on the shared historical context and the load-regime event.

\emph{MCQ subtask:}
\begin{itemize}[leftmargin=1.1em,itemsep=0pt,topsep=0pt,parsep=0pt,partopsep=0pt]
  \item Future vs. History
  (\textit{Higher / Lower / Similar / Uncertain}): \textbf{Similar}
  \item Volatility Change
  (\textit{increased / decreased / constant / Uncertain}): \textbf{Uncertain}
  \item Seasonality Shift
  (\textit{fixed / shifting / no / Uncertain}): \textbf{no}
\end{itemize}

\end{tcolorbox}
\caption{A full example instance (Causal Chambers) illustrating tasks T1--T4 in TemporalBench. All tiers share the same underlying physical-system time-series context, while T4 is additionally conditioned on a load-regime event.}
\label{fig:appendix_causal_chambers_one_example}
\end{figure}

To make the task design of TemporalBench concrete, we present one complete example instance for each domain in the appendix. Each example illustrates how a single time-series sample is transformed into four task tiers (T1--T4) under a unified abstraction, while preserving domain-specific characteristics.

Figure~\ref{fig:appendix_psml_one_example} shows an example from PSML, where the shared context consists of power load and meteorological covariates. The tasks range from basic time-series understanding (T1), to forecasting and qualitative judgments (T2), event-based reasoning (T3), and event-conditioned forecasting and decision-making (T4).

Figure~\ref{fig:appendix_mimic_one_example} presents a clinical example from MIMIC. The shared context contains multivariate physiological signals, with heart rate serving as a reference variable. In this setting, T3 and T4 are conditioned on a clinically meaningful event (fever), demonstrating how TemporalBench captures medically grounded temporal reasoning beyond pure forecasting.

Figure~\ref{fig:appendix_freshretailnet_one_example} illustrates a retail scenario from FreshRetailNet. The example highlights sparse and promotion-driven sales dynamics, where discount-related events play a central role in T3 and T4. Despite the domain shift, the task structure remains consistent with the other datasets.

Finally, Figure~\ref{fig:appendix_causal_chambers_one_example} provides an example from Causal Chambers, a controlled physical system. The shared context includes system measurements and operational variables, while T3 and T4 focus on reasoning under load-regime interventions, emphasizing causal and regime-based temporal comparisons.

Across all four examples, the same task tiers operate on different domains and data characteristics, illustrating that TemporalBench evaluates time-series understanding, forecasting, and event-driven reasoning within a single, coherent benchmark framework.

\section{Performance with different base LLMs}

\begin{table*}[t]
\centering
\footnotesize

\caption{Multi-choice (MCQ) task performance across datasets and agent frameworks under different backbone models.}
\label{tab:mcq_all_backbones}

\begin{subtable}[t]{\textwidth}
\centering
\caption{Base model: \texttt{claude-3.7-sonnet}}
\setlength{\tabcolsep}{2.8pt}
\renewcommand{\arraystretch}{1.2}

\begin{tabular}{c|c|cccc|cccc|cccc|cccc}
\toprule
\multirow{2}{*}{Agent} & Dataset
& \multicolumn{4}{c|}{FreshRetailNet}
& \multicolumn{4}{c|}{PSML}
& \multicolumn{4}{c|}{Causal Chambers}
& \multicolumn{4}{c}{MIMIC} \\
\cmidrule(lr){3-6}
\cmidrule(lr){7-10}
\cmidrule(lr){11-14}
\cmidrule(lr){15-18}
& Task
& T1 & T2 & T3 & T4
& T1 & T2 & T3 & T4
& T1 & T2 & T3 & T4
& T1 & T2 & T3 & T4 \\
\midrule

\multirow{2}{*}{Single LLM}
& SR
& 84.09\% & 100.00\% & 82.39\% & 100.00\%
& 98.00\% & 94.00\% & 100.00\% & 98.00\%
& 66.00\% & 100.00\% & 80.00\% & 78.00\%
& 100.00\% & 100.00\% & 100.00\% & 100.00\% \\
& ACC
& 50.00\% & 8.33\% & 67.88\% & 16.67\%
& 65.00\% & 19.18\% & 21.60\% & 33.33\%
& 8.00\% & 59.33\% & 28.40\% & 44.67\%
& 28.72\% & 36.17\% & 42.48\% & 32.62\% \\
\midrule

\multirow{2}{*}{\makecell[c]{TimeSeries\\Scientist}}
& SR
& 76.14\% & 100.00\% & 100.00\% & 100.00\%
& 72.50\% & 100.00\% & 99.60\% & 100.00\%
& 100.00\% & 100.00\% & 96.80\% & 100.00\%
& 77.13\% & 100.00\% & 100.00\% & 100.00\% \\
& ACC
& 26.14\% & 56.82\% & 3.98\% & 56.82\%
& 22.00\% & 26.67\% & 13.20\% & 27.33\%
& 28.67\% & 2.67\% & 23.60\% & 2.67\%
& 25.00\% & 23.40\% & 27.20\% & 23.40\% \\
\midrule

\multirow{2}{*}{AgentScope}
& SR
& 100.00\% & 79.55\% & 89.77\% & 100.00\%
& 100.00\% & 98.00\% & 100.00\% & 100.00\%
& 75.00\% & 94.00\% & 100.00\% & 100.00\%
& 100.00\% & 100.00\% & 100.00\% & 100.00\% \\
& ACC
& 50.57\% & 7.62\% & 35.80\% & 15.15\%
& 67.00\% & 19.73\% & 26.40\% & 30.67\%
& 11.33\% & 65.25\% & 44.80\% & 63.33\%
& 30.32\% & 33.33\% & 26.20\% & 27.66\% \\
\midrule

\multirow{2}{*}{MetaGPT}
& SR
& 100.00\% & 100.00\% & 82.95\% & 100.00\%
& 100.00\% & 86.00\% & 92.40\% & 96.00\%
& 75.00\% & 98.00\% & 100.00\% & 100.00\%
& 100.00\% & 100.00\% & 100.00\% & 100.00\% \\
& ACC
& 57.95\% & 10.61\% & 30.68\% & 12.12\%
& 65.00\% & 22.46\% & 22.80\% & 32.64\%
& 16.67\% & 64.00\% & 46.00\% & 63.33\%
& 31.38\% & 36.17\% & 23.59\% & 34.04\% \\
\midrule

\multirow{2}{*}{CAMEL}
& SR
& 100.00\% & 79.55\% & 89.77\% & 100.00\%
& 100.00\% & 98.00\% & 100.00\% & 100.00\%
& 75.00\% & 94.00\% & 100.00\% & 100.00\%
& 100.00\% & 100.00\% & 100.00\% & 100.00\% \\
& ACC
& 50.57\% & 7.62\% & 35.80\% & 15.15\%
& 67.00\% & 19.73\% & 26.40\% & 30.67\%
& 11.33\% & 65.25\% & 44.80\% & 63.33\%
& 30.32\% & 33.33\% & 26.20\% & 27.66\% \\
\bottomrule
\end{tabular}
\end{subtable}

\vspace{2pt}

\begin{subtable}[t]{\textwidth}
\centering
\caption{Base model: \texttt{gemini-2.5-flash}}
\setlength{\tabcolsep}{2.8pt}
\renewcommand{\arraystretch}{1.2}
\begin{tabular}{c|c|cccc|cccc|cccc|cccc}
\toprule
\multirow{2}{*}{Agent} & Dataset
& \multicolumn{4}{c|}{FreshRetailNet}
& \multicolumn{4}{c|}{PSML}
& \multicolumn{4}{c|}{Causal Chambers}
& \multicolumn{4}{c}{MIMIC} \\
\cmidrule(lr){3-6}
\cmidrule(lr){7-10}
\cmidrule(lr){11-14}
\cmidrule(lr){15-18}

& Task
& T1 & T2 & T3 & \multicolumn{1}{c|}{T4}
& T1 & T2 & T3 & \multicolumn{1}{c|}{T4}
& T1 & T2 & T3 & \multicolumn{1}{c|}{T4}
& T1 & T2 & T3 & T4 \\
\midrule

\multirow{2}{*}{Single LLM}
& SR
& 100.00\% & 0.00\% & 76.70\% & \multicolumn{1}{c|}{9.09\%}
& 100.00\% & 74.00\% & 100.00\% & \multicolumn{1}{c|}{82.00\%}
& 100.00\% & 100.00\% & 100.00\% & \multicolumn{1}{c|}{100.00\%}
& 100.00\% & 100.00\% & 94.66\% & 100.00\% \\
& ACC
& 19.89\% & 0.00\% & 16.56\% & \multicolumn{1}{c|}{1.52\%}
& 47.50\% & 12.00\% & 13.60\% & \multicolumn{1}{c|}{21.33\%}
& 11.33\% & 29.33\% & 50.40\% & \multicolumn{1}{c|}{14.67\%}
& 26.60\% & 20.57\% & 40.94\% & 21.99\% \\
\midrule

\multirow{2}{*}{\makecell[c]{TimeSeries\\Scientist}}
& SR
& 76.14\% & 100.00\% & 100.00\% & \multicolumn{1}{c|}{100.00\%}
& 72.50\% & 100.00\% & 99.60\% & \multicolumn{1}{c|}{100.00\%}
& 100.00\% & 100.00\% & 96.80\% & \multicolumn{1}{c|}{100.00\%}
& 77.13\% & 100.00\% & 100.00\% & 100.00\% \\
& ACC
& 26.14\% & 56.82\% & 3.98\% & \multicolumn{1}{c|}{56.82\%}
& 22.00\% & 26.67\% & 13.20\% & \multicolumn{1}{c|}{27.33\%}
& 28.67\% & 2.67\% & 23.60\% & \multicolumn{1}{c|}{2.67\%}
& 25.00\% & 23.40\% & 27.20\% & 23.40\% \\
\midrule

\multirow{2}{*}{AgentScope}
& SR
& 100.00\% & 72.73\% & 85.80\% & \multicolumn{1}{c|}{59.09\%}
& 100.00\% & 70.00\% & 98.40\% & \multicolumn{1}{c|}{68.00\%}
& 75.00\% & 100.00\% & 100.00\% & \multicolumn{1}{c|}{98.00\%}
& 100.00\% & 100.00\% & 100.00\% & 100.00\% \\
& ACC
& 27.27\% & 4.17\% & 26.70\% & \multicolumn{1}{c|}{15.38\%}
& 48.50\% & 23.81\% & 14.80\% & \multicolumn{1}{c|}{28.43\%}
& 12.00\% & 30.67\% & 38.40\% & \multicolumn{1}{c|}{29.93\%}
& 28.19\% & 18.44\% & 23.76\% & 24.82\% \\
\midrule

\multirow{2}{*}{MetaGPT}
& SR
& 100.00\% & 40.91\% & 71.59\% & \multicolumn{1}{c|}{15.91\%}
& 100.00\% & 82.00\% & 99.60\% & \multicolumn{1}{c|}{70.00\%}
& 75.00\% & 96.00\% & 100.00\% & \multicolumn{1}{c|}{100.00\%}
& 100.00\% & 100.00\% & 95.82\% & 100.00\% \\
& ACC
& 18.18\% & 7.41\% & 20.39\% & \multicolumn{1}{c|}{4.76\%}
& 47.50\% & 16.26\% & 15.20\% & \multicolumn{1}{c|}{23.81\%}
& 14.67\% & 30.56\% & 37.20\% & \multicolumn{1}{c|}{22.00\%}
& 27.13\% & 16.31\% & 23.71\% & 19.15\% \\
\midrule

\multirow{2}{*}{CAMEL}
& SR
& 100.00\% & 15.91\% & 85.23\% & \multicolumn{1}{c|}{9.09\%}
& 100.00\% & 76.00\% & 99.60\% & \multicolumn{1}{c|}{80.00\%}
& 75.00\% & 100.00\% & 100.00\% & \multicolumn{1}{c|}{94.00\%}
& 100.00\% & 100.00\% & 100.00\% & 100.00\% \\
& ACC
& 25.57\% & 9.52\% & 25.00\% & \multicolumn{1}{c|}{41.67\%}
& 48.00\% & 20.18\% & 10.40\% & \multicolumn{1}{c|}{24.17\%}
& 16.67\% & 28.67\% & 45.60\% & \multicolumn{1}{c|}{29.79\%}
& 23.94\% & 20.57\% & 24.61\% & 20.57\% \\
\bottomrule
\end{tabular}

\end{subtable}

\vspace{2pt}

\begin{subtable}[t]{\textwidth}
\centering
\caption{Base model: \texttt{deepseek-chat}}
\setlength{\tabcolsep}{2.8pt}
\renewcommand{\arraystretch}{1.2}

\begin{tabular}{c|c|cccc|cccc|cccc|cccc}
\toprule
\multirow{2}{*}{Agent} & Dataset
& \multicolumn{4}{c|}{FreshRetailNet}
& \multicolumn{4}{c|}{PSML}
& \multicolumn{4}{c|}{Causal Chambers}
& \multicolumn{4}{c}{MIMIC} \\
\cmidrule(lr){3-6}
\cmidrule(lr){7-10}
\cmidrule(lr){11-14}
\cmidrule(lr){15-18}

& Task
& T1 & T2 & T3 & \multicolumn{1}{c|}{T4}
& T1 & T2 & T3 & \multicolumn{1}{c|}{T4}
& T1 & T2 & T3 & \multicolumn{1}{c|}{T4}
& T1 & T2 & T3 & T4 \\
\midrule

\multirow{2}{*}{Single LLM}
& SR
& 100.00\% & 100.00\% & 82.39\% & \multicolumn{1}{c|}{100.00\%}
& 100.00\% & 100.00\% & 100.00\% & \multicolumn{1}{c|}{100.00\%}
& 100.00\% & 100.00\% & 100.00\% & \multicolumn{1}{c|}{100.00\%}
& 100.00\% & 100.00\% & 100.00\% & 100.00\% \\
& ACC
& 68.18\% & 56.82\% & 16.85\% & \multicolumn{1}{c|}{26.52\%}
& 74.00\% & 24.67\% & 34.80\% & \multicolumn{1}{c|}{36.00\%}
& 15.33\% & 19.33\% & 44.40\% & \multicolumn{1}{c|}{29.33\%}
& 33.00\% & 22.67\% & 39.14\% & 22.67\% \\
\midrule

\multirow{2}{*}{\makecell[c]{TimeSeries\\Scientist}}
& SR
& 100.00\% & 100.00\% & 100.00\% & \multicolumn{1}{c|}{100.00\%}
& 100.00\% & 100.00\% & 100.00\% & \multicolumn{1}{c|}{100.00\%}
& 100.00\% & 100.00\% & 100.00\% & \multicolumn{1}{c|}{100.00\%}
& 100.00\% & 100.00\% & 100.00\% & 100.00\% \\
& ACC
& 36.36\% & 56.82\% & 34.09\% & \multicolumn{1}{c|}{56.82\%}
& 38.50\% & 26.67\% & 22.80\% & \multicolumn{1}{c|}{27.33\%}
& 28.67\% & 2.67\% & 29.60\% & \multicolumn{1}{c|}{2.67\%}
& 15.43\% & 23.40\% & 25.94\% & 23.40\% \\
\midrule

\multirow{2}{*}{AgentScope}
& SR
& 100.00\% & 100.00\% & 96.02\% & \multicolumn{1}{c|}{100.00\%}
& 100.00\% & 78.00\% & 100.00\% & \multicolumn{1}{c|}{100.00\%}
& 75.00\% & 100.00\% & 100.00\% & \multicolumn{1}{c|}{100.00\%}
& 100.00\% & 100.00\% & 100.00\% & 100.00\% \\
& ACC
& 70.45\% & 34.85\% & 5.68\% & \multicolumn{1}{c|}{42.42\%}
& 68.50\% & 20.51\% & 24.80\% & \multicolumn{1}{c|}{31.33\%}
& 16.00\% & 47.33\% & 46.40\% & \multicolumn{1}{c|}{42.00\%}
& 32.98\% & 22.70\% & 25.71\% & 24.82\% \\
\midrule

\multirow{2}{*}{MetaGPT}
& SR
& 100.00\% & 100.00\% & 97.16\% & \multicolumn{1}{c|}{100.00\%}
& 100.00\% & 98.00\% & 100.00\% & \multicolumn{1}{c|}{100.00\%}
& 75.00\% & 100.00\% & 100.00\% & \multicolumn{1}{c|}{100.00\%}
& 100.00\% & 100.00\% & 100.00\% & 100.00\% \\
& ACC
& 69.89\% & 41.67\% & 5.11\% & \multicolumn{1}{c|}{37.12\%}
& 71.50\% & 19.73\% & 28.00\% & \multicolumn{1}{c|}{30.67\%}
& 15.33\% & 22.00\% & 46.40\% & \multicolumn{1}{c|}{40.00\%}
& 32.98\% & 22.70\% & 25.79\% & 30.50\% \\
\midrule

\multirow{2}{*}{CAMEL}
& SR
& 100.00\% & 100.00\% & 90.91\% & \multicolumn{1}{c|}{100.00\%}
& 100.00\% & 74.00\% & 100.00\% & \multicolumn{1}{c|}{100.00\%}
& 75.00\% & 100.00\% & 100.00\% & \multicolumn{1}{c|}{100.00\%}
& 100.00\% & 100.00\% & 100.00\% & 100.00\% \\
& ACC
& 67.05\% & 8.33\% & 19.89\% & \multicolumn{1}{c|}{29.55\%}
& 65.00\% & 18.02\% & 25.20\% & \multicolumn{1}{c|}{30.67\%}
& 12.67\% & 65.33\% & 46.80\% & \multicolumn{1}{c|}{48.00\%}
& 31.38\% & 24.82\% & 25.61\% & 31.21\% \\
\bottomrule
\end{tabular}
\end{subtable}

\vspace{2pt}

\begin{subtable}[t]{\textwidth}
\centering
\caption{Base model: \texttt{qwen-plus}}
\setlength{\tabcolsep}{2.8pt}
\renewcommand{\arraystretch}{1.2}

\begin{tabular}{c|c|cccc|cccc|cccc|cccc}
\toprule
\multirow{2}{*}{Agent} & Dataset
& \multicolumn{4}{c|}{FreshRetailNet}
& \multicolumn{4}{c|}{PSML}
& \multicolumn{4}{c|}{Causal Chambers}
& \multicolumn{4}{c}{MIMIC} \\
\cmidrule(lr){3-6}
\cmidrule(lr){7-10}
\cmidrule(lr){11-14}
\cmidrule(lr){15-18}

& Task
& T1 & T2 & T3 & \multicolumn{1}{c|}{T4}
& T1 & T2 & T3 & \multicolumn{1}{c|}{T4}
& T1 & T2 & T3 & \multicolumn{1}{c|}{T4}
& T1 & T2 & T3 & T4 \\
\midrule

\multirow{2}{*}{Single LLM}
& SR
& 100.00\% & 100.00\% & 64.20\% & \multicolumn{1}{c|}{70.45\%}
& 100.00\% & 54.00\% & 100.00\% & \multicolumn{1}{c|}{26.00\%}
& 100.00\% & 100.00\% & 100.00\% & \multicolumn{1}{c|}{100.00\%}
& 100.00\% & 100.00\% & 100.00\% & 100.00\% \\
& ACC
& 60.80\% & 11.36\% & 7.30\% & \multicolumn{1}{c|}{9.85\%}
& 49.00\% & 10.67\% & 23.20\% & \multicolumn{1}{c|}{8.00\%}
& 20.00\% & 18.00\% & 50.40\% & \multicolumn{1}{c|}{35.33\%}
& 43.50\% & 27.33\% & 40.84\% & 35.33\% \\
\midrule

\multirow{2}{*}{\makecell[c]{TimeSeries\\Scientist}}
& SR
& 17.61\% & 100.00\% & 100.00\% & \multicolumn{1}{c|}{100.00\%}
& 18.00\% & 100.00\% & 100.00\% & \multicolumn{1}{c|}{100.00\%}
& 100.00\% & 100.00\% & 75.20\% & \multicolumn{1}{c|}{100.00\%}
& 22.87\% & 100.00\% & 99.58\% & 100.00\% \\
& ACC
& 7.95\% & 56.82\% & 23.30\% & \multicolumn{1}{c|}{56.82\%}
& 12.50\% & 26.67\% & 14.00\% & \multicolumn{1}{c|}{27.33\%}
& 28.67\% & 2.67\% & 20.80\% & \multicolumn{1}{c|}{2.67\%}
& 5.85\% & 23.40\% & 28.45\% & 23.40\% \\
\midrule

\multirow{2}{*}{AgentScope}
& SR
& 100.00\% & 100.00\% & 100.00\% & \multicolumn{1}{c|}{90.91\%}
& 100.00\% & 84.00\% & 100.00\% & \multicolumn{1}{c|}{78.00\%}
& 75.00\% & 100.00\% & 100.00\% & \multicolumn{1}{c|}{100.00\%}
& 100.00\% & 100.00\% & 100.00\% & 100.00\% \\
& ACC
& 58.52\% & 7.58\% & 39.77\% & \multicolumn{1}{c|}{8.33\%}
& 54.50\% & 19.05\% & 27.60\% & \multicolumn{1}{c|}{26.50\%}
& 24.00\% & 42.67\% & 49.60\% & \multicolumn{1}{c|}{42.00\%}
& 44.15\% & 25.53\% & 28.24\% & 29.79\% \\
\midrule

\multirow{2}{*}{MetaGPT}
& SR
& 100.00\% & 22.73\% & 100.00\% & \multicolumn{1}{c|}{56.82\%}
& 100.00\% & 86.00\% & 100.00\% & \multicolumn{1}{c|}{88.00\%}
& 75.00\% & 100.00\% & 100.00\% & \multicolumn{1}{c|}{96.00\%}
& 100.00\% & 100.00\% & 100.00\% & 97.87\% \\
& ACC
& 65.34\% & 16.67\% & 40.34\% & \multicolumn{1}{c|}{10.67\%}
& 53.50\% & 18.60\% & 29.20\% & \multicolumn{1}{c|}{28.03\%}
& 26.00\% & 34.00\% & 49.20\% & \multicolumn{1}{c|}{39.58\%}
& 47.34\% & 26.24\% &26.95\%  & 31.88\% \\
\midrule

\multirow{2}{*}{CAMEL}
& SR
& 100.00\% & 100.00\% & 15.91\% & \multicolumn{1}{c|}{100.00\%}
& 100.00\% & 75.00\% & 100.00\% & \multicolumn{1}{c|}{75.00\%}
& 74.00\% & 100.00\% & 100.00\% & \multicolumn{1}{c|}{100.00\%}
& 100.00\% & 100.00\% & 100.00\% & 100.00\% \\
& ACC
& 64.20\% & 12.88\% & 35.80\% & \multicolumn{1}{c|}{12.88\%}
& 75.00\% & 44.44\% & 32.00\% & \multicolumn{1}{c|}{44.44\%}
& 22.67\% & 33.33\% & 48.40\% & \multicolumn{1}{c|}{42.67\%}
& 41.49\% & 25.53\% & 27.62\% & 34.04\% \\
\bottomrule
\end{tabular}
\end{subtable}

\end{table*}

\begin{table*}[t]
\centering
\footnotesize
\caption{Forecasting performance on T2 and T4 tasks across datasets and agent frameworks under different backbone models.}
\label{tab:forecast_all_backbones}

\begin{subtable}[t]{\textwidth}
\centering
\caption{Base model: \texttt{claude-3.7-sonnet}}
\setlength{\tabcolsep}{12pt}
\renewcommand{\arraystretch}{0.8}

\begin{tabular}{c|c|cc|cc|cc!{\vrule width 1.2pt}cc}
\toprule
\multirow{2}{*}{Agent} & \multirow{2}{*}{Metric}
& \multicolumn{2}{c|}{FreshRetailNet}
& \multicolumn{2}{c|}{PSML}
& \multicolumn{2}{c!{\vrule width 1.2pt}}{Causal Chambers}
& \multicolumn{2}{c}{MIMIC} \\
\cmidrule(lr){3-4}
\cmidrule(lr){5-6}
\cmidrule(lr){7-8}
\cmidrule(lr){9-10}

&
& T2 & T4
& T2 & T4
& T2 & T4
& T2 & T4 \\
\midrule

\multirow{3}{*}{Single LLM}
& SR
& 61.36\% & 47.73\%
& 34.00\% & 40.00\%
& 48.00\% & 46.00\%
& 0.00\% & 97.87\% \\
& MAE
& 0.11 & 0.14
& 0.26 & 0.29
& 2.31 & 2.70
&  & 12.14 \\
& sMAPE
& 1.25 & 1.22
& 0.28 & 0.30
& 2.40E{-}05 & 2.80E{-}05
&  & 0.47 \\
\midrule

\multirow{3}{*}{\makecell[c]{TimeSeries\\Scientist}}
& SR
& 36.36\% & 43.18\%
& 22.00\% & 14.00\%
& 54.00\% & 30.00\%
& 89.36\% & 89.36\% \\
& MAE
& 0.14 & 0.16
& 0.26 & 0.26
& 2.32 & 2.57
& 9.75 & 10.54 \\
& sMAPE
& 1.29 & 1.21
& 0.29 & 0.24
& 2.41E{-}05 & 2.68E{-}05
& 0.34 & 0.38 \\
\midrule

\multirow{3}{*}{AgentScope}
& SR
& 65.91\% & 88.64\%
& 30.00\% & 34.00\%
& 62.00\% & 32.00\%
& 97.87\% & 95.74\% \\
& MAE
& 0.13 & 0.11
& 0.27 & 0.24
& 2.16 & 2.79
& 9.86 & 13.45 \\
& sMAPE
& 121.00 & 121.96
& 26.92 & 20.07
& 2.25E{-}03 & 2.90E{-}03
& 0.41 & 0.48 \\
\midrule

\multirow{3}{*}{MetaGPT}
& SR
& 29.55\% & 70.45\%
& 14.00\% & 26.00\%
& 50.00\% & 36.00\%
& 97.87\% & 97.87\% \\
& MAE
& 0.12 & 0.14
& 0.16 & 0.24
& 2.07 & 2.87
& 9.90 & 11.36 \\
& sMAPE
& 126.50 & 120.66
& 16.88 & 26.33
& 2.15E{-}03 & 2.99E{-}03
& 0.41 & 0.49 \\
\midrule

\multirow{3}{*}{CAMEL}
& SR
& 65.91\% & 88.64\%
& 30.00\% & 34.00\%
& 62.00\% & 32.00\%
& 97.87\% & 95.74\% \\
& MAE
& 0.13 & 0.11
& 0.27 & 0.24
& 2.16 & 2.79
& 9.86 & 13.45 \\
& sMAPE
& 121.00 & 121.96
& 26.92 & 20.07
& 2.25E{-}03 & 2.90E{-}03
& 0.41 & 0.48 \\
\bottomrule
\end{tabular}
\end{subtable}

\vspace{2pt}

\begin{subtable}[t]{\textwidth}
\centering
\caption{Base model: \texttt{gemini-2.5-flash}}
\setlength{\tabcolsep}{12pt}
\renewcommand{\arraystretch}{0.8}

\begin{tabular}{c|c|cc|cc|cc!{\vrule width 1.2pt}cc}
\toprule
\multirow{2}{*}{Agent} & \multirow{2}{*}{Metric}
& \multicolumn{2}{c|}{FreshRetailNet}
& \multicolumn{2}{c|}{PSML}
& \multicolumn{2}{c!{\vrule width 1.2pt}}{Causal Chambers}
& \multicolumn{2}{c}{MIMIC} \\
\cmidrule(lr){3-4}
\cmidrule(lr){5-6}
\cmidrule(lr){7-8}
\cmidrule(lr){9-10}

&
& T2 & T4
& T2 & T4
& T2 & T4
& T2 & T4 \\
\midrule

\multirow{3}{*}{Single LLM}
& SR
& 0.00\% & 2.27\%
& 0.00\% & 4.00\%
& 36.00\% & 38.00\%
& 0.00\% & 89.36\% \\
& MAE
&  & 0.06
&  & 0.11
& 2.18 & 2.90
&  & 11.80 \\
& sMAPE
&  & 1.29
&  & 0.16
& 2.26E{-}05 & 3.02E{-}05
&  & 0.38 \\
\midrule

\multirow{3}{*}{\makecell[c]{TimeSeries\\Scientist}}
& SR
& 36.36\% & 43.18\%
& 22.00\% & 14.00\%
& 54.00\% & 30.00\%
& 89.36\% & 89.36\% \\
& MAE
& 0.14 & 0.16
& 0.26 & 0.26
& 2.32 & 2.57
& 9.75 & 10.54 \\
& sMAPE
& 1.29 & 1.21
& 0.29 & 0.24
& 2.41E{-}05 & 2.68E{-}05
& 0.34 & 0.38 \\
\midrule

\multirow{3}{*}{AgentScope}
& SR
& 0.00\% & 2.33\%
& 0.00\% & 2.04\%
& 30.19\% & 28.30\%
& 75.47\% & 82.35\% \\
& MAE
&  & 0.15
&  & 0.14
& 2.32 & 2.20
& 11.60 & 12.06 \\
& sMAPE
&  & 143.19
&  & 16.93
& 2.41E{-}03 & 2.29E{-}03
& 0.41 & 0.36 \\
\midrule

\multirow{3}{*}{MetaGPT}
& SR
& 0.00\% & 4.55\%
& 4.00\% & 6.00\%
& 50.00\% & 42.00\%
& 76.60\% & 89.36\% \\
& MAE
&  & 0.39
& 0.24 & 0.20
& 2.48 & 2.55
& 11.26 & 10.93 \\
& sMAPE
&  & 124.81
& 24.13 & 21.58
& 2.58E{-}03 & 2.65E{-}03
& 0.40 & 0.35 \\
\midrule

\multirow{3}{*}{CAMEL}
& SR
& 0.00\% & 0.00\%
& 2.00\% & 4.00\%
& 34.00\% & 46.00\%
& 53.19\% & 87.23\% \\
& MAE
&  & 0.16
& 0.16 & 0.16
& 2.34 & 2.62
& 11.08 & 11.58 \\
& sMAPE
&  & 22.22
& 19.38 & 19.38
& 2.43E{-}03 & 2.73E{-}03
& 0.39 & 0.36 \\
\bottomrule
\end{tabular}
\end{subtable}

\vspace{2pt}

\begin{subtable}[t]{\textwidth}
\centering
\caption{Base model: \texttt{deepseek-chat}}
\setlength{\tabcolsep}{12pt}
\renewcommand{\arraystretch}{0.8}

\begin{tabular}{c|c|cc|cc|cc!{\vrule width 1.2pt}cc}
\toprule
\multirow{2}{*}{Agent} & \multirow{2}{*}{Metric}
& \multicolumn{2}{c|}{FreshRetailNet}
& \multicolumn{2}{c|}{PSML}
& \multicolumn{2}{c!{\vrule width 1.2pt}}{Causal Chambers}
& \multicolumn{2}{c}{MIMIC} \\
\cmidrule(lr){3-4}
\cmidrule(lr){5-6}
\cmidrule(lr){7-8}
\cmidrule(lr){9-10}

&
& T2 & T4
& T2 & T4
& T2 & T4
& T2 & T4 \\
\midrule

\multirow{3}{*}{Single LLM}
& SR
& 0.00\% & 2.27\%
& 20.00\% & 46.00\%
& 52.00\% & 48.00\%
& 0.00\% & 98.00\% \\
& MAE
&  & 0.10
& 0.29 & 0.38
& 1.97 & 2.53
&  & 12.12 \\
& sMAPE
&  & 0.97
& 0.27 & 0.28
& 2.05E{-}05 & 2.63E{-}05
&  & 0.48 \\
\midrule

\multirow{3}{*}{\makecell[c]{TimeSeries\\Scientist}}
& SR
& 34.09\% & 25.00\%
& 68.00\% & 60.00\%
& 64.00\% & 40.00\%
& 89.36\% & 89.36\% \\
& MAE
& 0.17 & 0.12
& 0.34 & 0.34
& 2.11 & 2.51
& 10.70 & 0.34 \\
& sMAPE
& 1.30 & 1.25
& 0.32 & 0.27
& 2.19E{-}05 & 2.61E{-}05
& 12.34 & 0.45 \\
\midrule

\multirow{3}{*}{AgentScope}
& SR
& 27.54\% & 1.15\%
& 12.36\% & 25.00\%
& 58.00\% & 40.00\%
& 97.87\% & 95.74\% \\
& MAE
& 0.10 & 0.07
& 0.29 & 0.36
& 2.43 & 2.72
& 10.22 & 11.43 \\
& sMAPE
& 124.06 & 129.35
& 27.39 & 27.94
& 2.52E{-}03 & 2.82E{-}03
& 0.46 & 0.39 \\
\midrule

\multirow{3}{*}{MetaGPT}
& SR
& 38.64\% & 0.00\%
& 22.00\% & 38.00\%
& 58.00\% & 44.00\%
& 97.87\% & 97.87\% \\
& MAE
&  & 0.09
& 0.31 & 0.35
& 2.33 & 2.70
& 10.29 & 10.88 \\
& sMAPE
&  & 125.21
& 25.15 & 29.31
& 2.42E{-}03 & 2.81E{-}03
& 0.41 & 0.41 \\
\midrule

\multirow{3}{*}{CAMEL}
& SR
& 45.45\% & 18.18\%
& 20.00\% & 30.00\%
& 56.00\% & 38.00\%
& 97.87\% & 97.87\% \\
& MAE
& 0.11 & 0.22
& 0.28 & 0.39
& 2.37 & 2.55
& 11.36 & 11.32 \\
& sMAPE
& 129.65 & 132.99
& 25.78 & 32.22
& 2.46E{-}03 & 2.65E{-}03
& 0.47 & 0.46 \\
\bottomrule
\end{tabular}
\end{subtable}

\vspace{2pt}

\begin{subtable}[t]{\textwidth}
\centering
\caption{Base model: \texttt{qwen-plus}}
\setlength{\tabcolsep}{12pt}
\renewcommand{\arraystretch}{0.8}

\begin{tabular}{c|c|cc|cc|cc!{\vrule width 1.2pt}cc}
\toprule
\multirow{2}{*}{Agent} & \multirow{2}{*}{Metric}
& \multicolumn{2}{c|}{FreshRetailNet}
& \multicolumn{2}{c|}{PSML}
& \multicolumn{2}{c!{\vrule width 1.2pt}}{Causal Chambers}
& \multicolumn{2}{c}{MIMIC} \\
\cmidrule(lr){3-4}
\cmidrule(lr){5-6}
\cmidrule(lr){7-8}
\cmidrule(lr){9-10}

&
& T2 & T4
& T2 & T4
& T2 & T4
& T2 & T4 \\
\midrule

\multirow{3}{*}{Single LLM}
& SR
& 0.00\% & 4.55\%
& 6.00\% & 12.00\%
& 46.00\% & 46.00\%
& 0.00\% & 92.00\% \\
& MAE
&  & 0.32
& 0.32 & 0.54
& 2.17 & 2.83
&  & 17.02 \\
& sMAPE
&  & 1.32
& 0.27 & 0.34
& 2.26E{-}05 & 2.95E{-}05
&  & 0.54 \\
\midrule

\multirow{3}{*}{\makecell[c]{TimeSeries\\Scientist}}
& SR
& 4.55\% & 6.82\%
& 0.00\% & 16.00\%
& 46.00\% & 44.00\%
& 82.98\% & 80.85\% \\
& MAE
& 0.06 & 0.28
&  & 0.39
& 2.43 & 2.46
& 15.08 & 18.80 \\
& sMAPE
& 1.30 & 1.34
&  & 0.32
& 2.52E{-}05 & 2.55E{-}05
& 0.42 & 0.58 \\
\midrule

\multirow{3}{*}{AgentScope}
& SR
& 1.15\% & 8.64\%
& 6.38\% & 6.38\%
& 39.22\% & 35.29\%
& 54.10\% & 89.80\% \\
& MAE
& 0.23 & 0.27
& 0.31 & 0.66
& 2.39 & 2.77
& 15.67 & 18.65 \\
& sMAPE
& 135.87 & 136.62
& 29.69 & 40.12
& 2.48E{-}03 & 2.88E{-}03
& 0.52 & 0.53 \\
\midrule

\multirow{3}{*}{MetaGPT}
& SR
& 0.00\% & 6.82\%
& 16.00\% & 12.00\%
& 42.00\% & 32.00\%
& 59.57\% & 100.00\% \\
& MAE
&  & 0.35
& 0.47 & 0.34
& 2.13 & 2.49
& 15.97 & 17.68 \\
& sMAPE
&  & 134.81
& 35.39 & 28.81
& 2.22E{-}03 & 2.59E{-}03
& 0.46 & 0.58 \\
\midrule

\multirow{3}{*}{CAMEL}
& SR
& 2.27\% & 100.00\%
& 20.00\% & 40.00\%
& 44.00\% & 40.00\%
& 72.34\% & 95.74\% \\
& MAE
& 0.16 & 0.23
& 0.08 & 0.32
& 2.15 & 2.54
& 14.62 & 18.26 \\
& sMAPE
& 125.89 & 120.09
& 7.58 & 24.99
& 2.24E{-}03 & 2.64E{-}03
& 0.50 & 0.56 \\
\bottomrule
\end{tabular}
\end{subtable}

\end{table*}

This section analyzes how agent performance varies when different large language models are used as the backbone. We compare \texttt{claude-3.7\allowbreak-sonnet}, \texttt{gemini-2.5\allowbreak-flash}, \texttt{deepseek\allowbreak-chat} and \texttt{qwen\allowbreak-plus}, and relate their behaviors to the trends observed in the main text where \texttt{gpt-4o} is used as the base model.

Across all backbone models, several consistent patterns emerge. First, agent-based frameworks substantially outperform the single-LLM baseline on multi-choice tasks (T1 and T3), particularly on reasoning-intensive subtasks. This mirrors the observations under \texttt{gpt-4o}, indicating that the gains from agent decomposition and tool-mediated reasoning are largely model-agnostic. Even when the underlying LLM is weaker or less stable, agents such as TimeSeriesScientist, AgentScope, MetaGPT, and CAMEL maintain high execution success rates, suggesting that structured workflows effectively mitigate backbone limitations.

Second, the accuracy gap between single LLMs and agent frameworks widens as tasks become more reasoning-heavy. Under all base models, T3 exhibits the largest performance variance, with single LLMs often collapsing to low accuracy despite high success rates, while agent frameworks retain meaningful gains. This trend is consistent with the \texttt{gpt-4o} results in the main paper and confirms that T3 exposes intrinsic reasoning deficiencies that cannot be resolved by prompt-only approaches.

Third, forecasting tasks (T2 and T4) show stronger dependence on the backbone model. Compared to \texttt{gpt-4o}, alternative base LLMs generally yield lower success rates and higher error metrics, especially on PSML and Causal Chambers. Nevertheless, agent frameworks still demonstrate more robust behavior than single LLMs, with reduced failure rates and more stable error profiles. This suggests that while numerical forecasting quality is sensitive to the backbone model, agent coordination improves reliability even when absolute accuracy remains constrained by model capacity.

Finally, differences among base LLMs primarily affect the absolute performance level rather than the relative ranking of agent frameworks. Models with stronger instruction-following and reasoning capabilities, such as \texttt{claude-3.7-sonnet}, exhibit patterns most similar to \texttt{gpt-4o}, whereas lighter or more compressed models (e.g., \texttt{gemini-2.5-flash}) show larger drops on forecasting tasks. Importantly, no base model eliminates the advantages of agent-based designs, reinforcing the central claim that temporal reasoning benefits from explicit structure beyond raw model scale.

Overall, these results demonstrate that the conclusions drawn in the main text using \texttt{gpt-4o} generalize across a diverse set of backbone LLMs. Agent-based temporal reasoning consistently improves robustness and accuracy, while the choice of base model primarily influences the ceiling rather than the existence of these gains.

\end{document}